\documentclass[journal]{IEEEtran}

\IEEEoverridecommandlockouts
\usepackage{hyperref}
\usepackage{cite}
\usepackage{lscape}
\usepackage{diagbox}
\usepackage{color,soul}
\usepackage{amsmath,amssymb,amsfonts}
\usepackage{dsfont}
\usepackage{mathtools}

\usepackage{cancel}
\usepackage{textcomp}
\usepackage{algorithm}
\usepackage{algpseudocode}
\usepackage{multirow}
\usepackage{graphicx}
\usepackage{textcomp}
\usepackage{xcolor}
\usepackage{cite}
\usepackage{enumitem}
\usepackage{tikz}
\usepackage{booktabs}
\newcommand{\tikzcircle}[2][red,fill=red]{\tikz[baseline=-0.5ex]\draw[#1,radius=#2] (0,0) circle ;}
\newcommand{\bm}[1]{\mbox{\boldmath{$#1$}}}

\definecolor{dvl}{HTML}{CB00FF}
\definecolor{slam}{HTML}{18DBD3}
\definecolor{smooth}{HTML}{DB00BE}
\definecolor{odom}{HTML}{EBBF17}
\definecolor{true}{HTML}{00CC07}
\definecolor{n0}{HTML}{000004}
\definecolor{n1}{HTML}{320a5e}
\definecolor{n2}{HTML}{781c6d}
\definecolor{n3}{HTML}{bc3754}
\definecolor{n4}{HTML}{ed6925}
\definecolor{n5}{HTML}{fbb61a}

\def\BibTeX{{\rm B\kern-.05em{\sc i\kern-.025em b}\kern-.08em
    T\kern-.1667em\lower.7ex\hbox{E}\kern-.125emX}}
\begin{document}
\newcommand{\pcd}[1]{\prescript{#1}{}{\mathcal{P}}}
\newcommand{\point}[1]{\prescript{#1}{}{\bm{p}}}

\newcommand{\R}[0]{\mathbb{R}}  
\newcommand{\I}[0]{\bm{I}}  
\newcommand{\SE}[0]{\text{SE}}
\newcommand{\SO}[0]{\text{SO}}  

\newcommand{\Wf}[0]{\text{W}}
\newcommand{\Tf}[0]{\text{T}}
\newcommand{\Cf}[0]{\text{C}}

\newcommand{\rotM}[0]{\bm{R}}
\newcommand{\Rot}[2]{\prescript{#1}{#2}{\mathbf{R}}}
\newcommand{\RotHat}[2]{\prescript{#1}{#2}{\mathbf{\hat{R}}}}
\newcommand{\Tran}[3]{\prescript{#1}{}{\bm{t}_{#2/#3}}}
\newcommand{\TranHat}[3]{\prescript{#1}{}{\hat{\bm{t}}_{#2/#3}}}
\newcommand{\TranStar}[3]{\prescript{#1}{}{\bm{t}^*_{#2/#3}}}
\newcommand{\tran}[3]{\prescript{#1}{}{\mathbf{t}_{#2/#3}}}
\newcommand{\Unit}[3]{\prescript{#1}{}{\hat{\bm{u}}_{#2/#3}}}
\newcommand{\Vel}[3]{\prescript{#1}{}{\bm{v}_{#2/#3}}}
\newcommand{\vel}[3]{\prescript{#1}{}{\mathbf{v}_{#2/#3}}}
\newcommand{\Tfm}[2]{\prescript{#1}{#2}{\mathbf{T}}}
\newcommand{\TfmHat}[2]{\prescript{#1}{#2}{\hat{\mathbf{T}}}}
\newcommand{\TfmBar}[2]{\prescript{#1}{#2}{\bar{\mathbf{T}}}}
\newcommand{\TfmMat}[2]{\begin{bmatrix} #1 & #2 \\[0.3em] \bm{0}^\top & 1 \\[0.3em]\end{bmatrix}}
\newcommand{\Skew}[1]{{#1}^\wedge}
\newcommand{\RefCov}[1]{\prescript{#1}{}{\bm \Sigma}}
\newcommand{\RefEta}[1]{\prescript{#1}{}{\bm \eta}}

\newcommand{\p}[0]{\bm{p}}
\newcommand{\bmell}[0]{\bm{\ell}}
\newcommand{\z}[0]{\bm{z}}

\newcommand{\calX}[0]{\mathcal{X}}

\newcommand{\barA}[0]{\bar{A}}
\newcommand{\barB}[0]{\bar{B}}

\newcommand{\chx}[0]{\bm{x}_{\text{C}}}
\newcommand{\tgtx}[0]{\bm{x}_{\text{T}}}
\newcommand{\ttgtx}[0]{\bm{x}_{\text{TT}}}
\newcommand{\tgth}[1]{\prescript{W}{}{\theta}_{#1}}
\newcommand{\tgtv}[1]{\prescript{T}{}{v}_{#1}}
\newcommand{\ttgtv}[1]{\prescript{T}{}{\bm v}_{#1}}
\newcommand{\ttgtw}[1]{\prescript{T}{}{\bm \omega}_{#1}}

\newcommand{\imuf}[0]{\phi_{\text{imu}}}
\newcommand{\odomf}[0]{\phi_{\text{odom}}}
\newcommand{\rbf}[0]{\phi_{\text{rb}}}
\newcommand{\velf}[0]{\phi_{\text{v}}}
\newcommand{\depthf}[0]{\phi_{\text{d}}}
\newcommand{\mmf}[0]{\phi_{\text{mm}}}
\newcommand{\priorf}[0]{\phi_{\text{p}}}
\newcommand{\priorfc}[0]{\phi_{\text{p}_{\text{C}}}}
\newcommand{\priorft}[0]{\phi_{\text{p}_{\text{T}}}}
\newcommand{\expf}[0]{\phi_{\text{exp}}}
\newcommand{\redf}[0]{\phi_{\text{red}}}
\newcommand{\optf}[0]{\phi_{\text{opt}}}
\newcommand{\tof}[0]{\phi_{\text{to}}}

\title{STERN: Simultaneous Trajectory Estimation and Relative Navigation for Autonomous Underwater Proximity Operations\\

}

\author{Aldo Ter\'an Espinoza, \IEEEmembership{Student Member, IEEE},
 Antonio Ter\'an Espinoza,
 John Folkesson, \IEEEmembership{Senior Member, IEEE},
 Clemens Deutsch,
 Niklas Rolleberg,
 Peter Sigray,
 Jakob Kuttenkeuler
\thanks{This work was supported by the Stiftelsen för Strategisk Forskning (SSF) through the Swedish Maritime Robotics Centre (SMaRC)(IRC15-0046).}
\thanks{Aldo Ter\'an Espinoza, Niklas Rolleberg, Clemens Deutsch, Peter Sigray, and Jakob Kuttenkeuler are with the Centre for Naval Architecture, KTH Royal Institute of Technology, 10044 Stockholm, Sweden. (e-mail: aldot@kth.se, nrol@kth.se, clemensd@kth.se, sigray@kth.se, jakob@kth.se)}
\thanks{Antonio Ter\'an Espinoza was with the Massachusetss Institue of Technology, Cambridge, MA 02142 USA. (e-mail: teran@mit.edu)}
\thanks{John Folkesson is with the Division of Robotics, Perception, and Learning, KTH Royal Institute of Technology, 10044 Stockholm, Sweden. (e-mail: johnf@kth.se)}
\thanks{© 2026 The Author(s). This work is licensed under a Creative Commons Attribution 4.0 International License (CC BY 4.0). The final version is available at https://doi.org/10.1109/JOE.2025.3624470}
}

\maketitle

\begin{abstract}
Due to the challenges regarding the limits of their endurance and autonomous capabilities, underwater docking for autonomous underwater vehicles (AUVs) has become a topic of interest for many academic and commercial applications. Herein, we take on the problem of relative navigation for the generalized version of the docking operation, which we address as proximity operations. Proximity operations typically involve only two actors, a chaser and a target. We leverage the similarities to proximity operations (prox-ops) from spacecraft robotic missions to frame the diverse docking scenarios with a set of phases the chaser undergoes on the way to its target. We emphasize the versatility on the use of factor graphs as a generalized representation to model the underlying simultaneous trajectory estimation and relative navigation (STERN) problem that arises with any prox-ops scenario, regardless of the sensor suite or the agents' dynamic constraints. To emphasize the flexibility of factor graphs as the modeling foundation for arbitrary underwater prox-ops, we compile a list of state-of-the-art research in the field and represent the different scenario using the same factor graph representation. We detail the procedure required to model, design, and implement factor graph-based estimators by addressing a long-distance acoustic homing scenario of an AUV to a moving mothership using datasets from simulated and real-world deployments; an analysis of these results is provided to shed light on the flexibility and limitations of the dynamic assumptions of the moving target. A description of our front- and back-end is also presented together with a timing breakdown of all processes to show its potential deployment on a real-time system. 
\end{abstract}

\begin{IEEEkeywords}
underwater docking, relative navigation, underwater proximity operations, factor graphs, state estimation, autonomous underwater vehicle navigation
\end{IEEEkeywords}

\section{Introduction}

\IEEEPARstart{U}{nderwater} missions play a critical role in many academic and commercial applications. For example,
academic institutions rely on oceanographic information for conducting a plethora of research~\cite{wynn_autonomous_2014, cowen_underwater_1997}, whereas the commercial sector (e.g., energy) requires underwater exploration and inspection of their subsea infrastructure to ensure safe and proper operations~\cite{krupinski_investigation_2008}. Traditionally, crewed missions are employed to carry out these underwater missions, if at all possible~\cite{gerovasileiou_three-dimensional_2013, rahn_diver_2015, kudo_overseas_2008}. While effective, these crewed missions are often prohibitively expensive and/or dangerous, especially given the hostile environment to which they are commonly subjected~\cite{zereik_challenges_2018}. Certain missions are beyond the scope of human operators~\cite{bellingham_arctic_2000, kaminski_12_2010}. 

\begin{figure}[t]
\centerline{\includegraphics[width=0.4\textwidth]{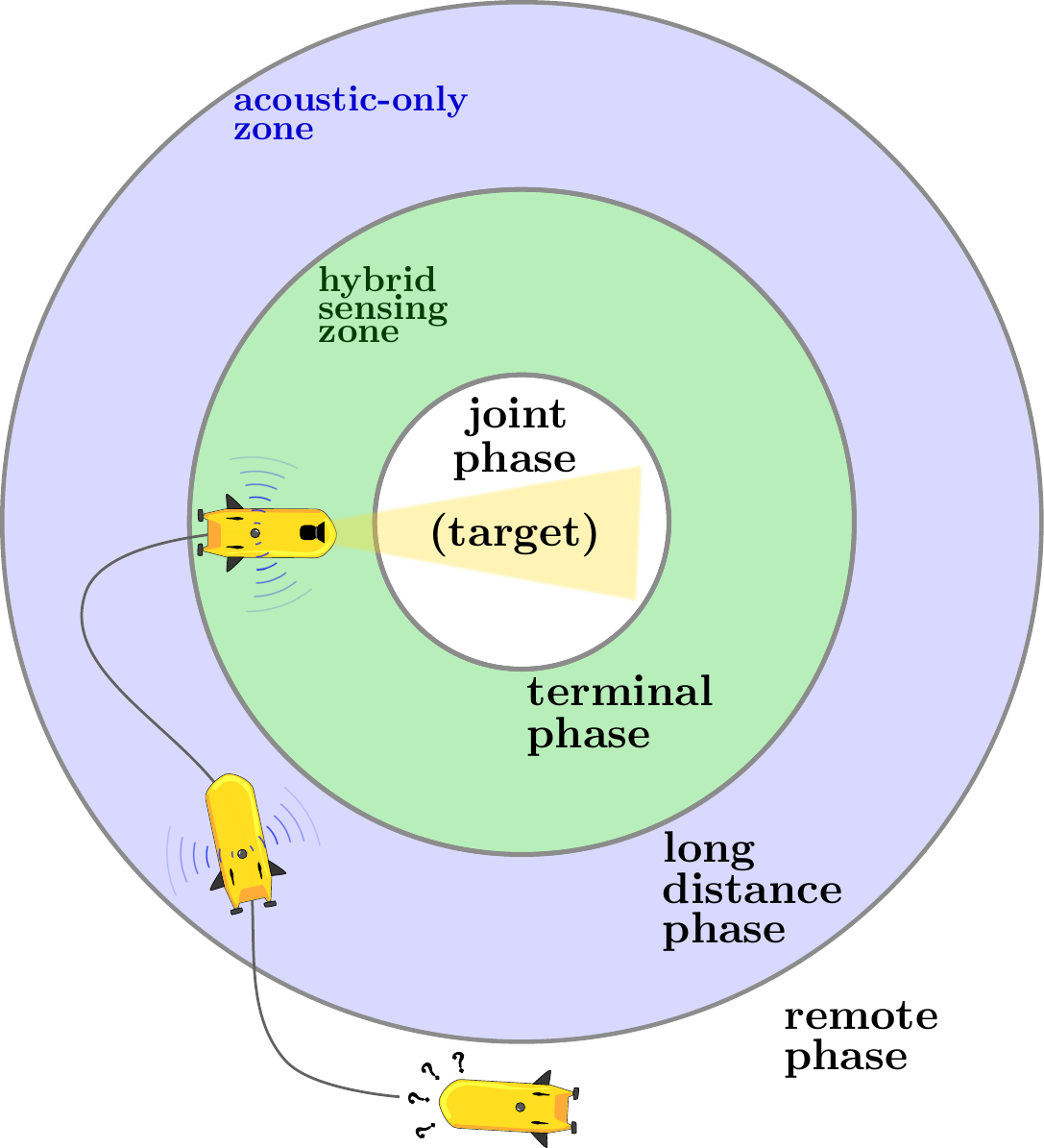}}
\caption{Definition of the prox-ops phases established to generalize any underwater missions consisting of a chaser and a target. The chaser is tasked with navigating through the phases using different sensor modalities, and the target is the objective the chaser must home into to execute some sort of joint operation.}
\label{fig:general_docking_phases}
\end{figure}

To overcome these limitations, robotic platforms such as remotely operated vehicles (ROVs) and autonomous underwater vehicles (AUVs) have come into the spotlight as safer and more affordable alternatives~\cite{wynn_autonomous_2014, mallios_toward_2016, zereik_challenges_2018}.
These robotic platforms have been shown to be quite effective at a broad swath of underwater missions involving intervention and manipulation~\cite{kim_underwater_2016, ridao_intervention_2015}, sampling and surveying~\cite{yoerger_autonomous_2007, huvenne_rovs_2018, williams_monitoring_2012}, and inspection~\cite{mai_subsea_2016} tasks, among many others.

Contrary to their autonomous counterparts (i.e., AUVs), ROVs suffer from major disadvantages that mostly stem from their reliance on a physical cable (a.k.a. an umbilical or a tether) for power and communication purposes. This cable significantly impairs the ROV's range and mobility capabilities~\cite{ridao_intervention_2015, mallios_toward_2016}.
On top of this, ROVs require constant supervision and active control by human operators, which further renders their deployment to be more costly and difficult~\cite{ridao_intervention_2015, trslic_vision_2020}.
In comparison, AUVs provide greater mission flexibility and accessibility by eliminating the inherent and aforementioned limitations of ROVs, enabling them to manipulate, inspect, and sample a more arbitrary set of targets throughout their deployments~\cite{ridao_intervention_2015}.

Naturally, AUVs come with their own set of challenges, most of them pertaining to the limits of their autonomous capabilities and endurance~\cite{yazdani_survey_2020}. 
Herein we are interested in target-based missions where an AUV is tasked to autonomously rendezvous or somehow interact with a specific target to accomplish a certain type of joint operation.
Examples of this type of mission can include docking with a moored charging station to replenish its batteries, homing into a mothership for recovery, or meeting with another AUV to share information.
While the nature of these missions can vary considerably, they all involve operations between two agents. 

To abstractly reason about this type of operations, it is convenient to borrow the proximity operations (prox-ops) term commonly used for spacecraft robotic missions, which describes target-based missions in terms of an actor---referred to as the chaser---that has to go through a series of phases to successfully reach its target~\cite{teran_espinoza_end--end_2019}.
Moreover, the use of this framework allows us to easily categorize and compare existing methods that aim at solving underwater prox-ops scenarios with different formulations and assumptions regarding the dynamics of the target.

In this work, we focus on underwater proximity operations between a single chaser and target agent.
One of the most important characteristics of an underwater prox-ops are the target dynamics, which we categorize as being either passive (i.e., static or free-floating) or dynamically active.
The more complex the dynamics of the target, the more difficult it becomes to estimate the evolution of its state throughout the different phases of the operation.
The more general scenario involves an active target and requires not only to localize the chasing AUV relative to the moving target, but also to simultaneously and explicitly reason about the pose and dynamics of the target.
The need to jointly estimate both chaser and target states motivates the use of modern probabilistic estimation methodologies such as simultaneous localization and mapping (SLAM) to successfully reach a solution~\cite{rosen_advances_2021}.
Simpler operations with static targets (such as docking scenarios involving fixed docking stations) can be reduced to a localization-only problem by assuming a negligible dynamics model for the target in the more general problem formulation. 

Current state-of-the-art research on underwater robotic proximity operations, which mainly falls under the umbrella of underwater docking, is mostly focused on developing and studying methods that solve very specific scenarios, sensor suites, and environmental dynamics---rendering these approaches hard to compare, to enhance or build upon, and in some cases to reproduce. Thus, herein, we propose the use of factor graphs to represent any underwater prox-ops scenario that relies on an explicit relative navigation module in order to be completed. Throughout the following sections, we support the claim of factor graphs being a general way of modeling any such scenario by shedding light into their intuitive structure and design procedure, and the different flavor of estimators that one can use to solve the underlying information-fusion problem. We identify and tackle one of the least researched prox-ops scenarios for the underwater domain, namely that of an AUV homing into a collaborative moving target by means of relative acoustic measurements, and thoroughly detail a basic solution to this scenario using our proposed framework.

Concisely, the main contributions of this work are as follows:
\begin{itemize}
    \item Provide a framework that borrows proven methodologies from robotic spacecraft proximity operations to represent the general information fusion problem intrinsic to underwater proximity operations.
    \item Emphasize the versatility and intuitive nature of modeling arbitrary underwater proximity operations using factor graphs by framing the identified related work within the Simultaneous Trajectory Estimation and Relative Navigation (STERN) framework.
    \item Detail the development and implementation of a basic solution to the long-distance acoustic homing to a moving target scenario as a STERN problem, while providing insights and detailing the limitations of the framework through a study on the performance of our STERN solution using simulated and real-world data.
\end{itemize}

The paper is divided into the following sections: Section \ref{sec:background} will introduce the background and description of a general prox-ops scenario and explain how factor graphs can be used to model the STERN problem, and use this framework to categorize the related work and establish the existing technical gap addressed throughout the article; Section \ref{sec:prop_approach} gives a thorough description of the proximity operation scenario at hand and carefully details the procedure for modeling and representing the STERN-based solution associated with it; the following Sections \ref{sec:simulations} and \ref{sec:results} go through the implementation, verification, and experimental validation of the proposed approach undertaken to corroborate its performance; last, Section \ref{sec:conclusions} discusses the key findings and possible future paths for further research.

\section{Background and Scenario Description}
\label{sec:background}

This section describes the general prox-ops scenario and the factor graph representation that we use as a framework to abstractly reason about any type of joint underwater close-proximity operation. We start by breaking things down into four fundamental phases. Subsequently, we use these phases to create a general scenario description that acts as common context for framing the existing related work using our proposed framework.

\subsection{Underwater Proximity Operation Phases}

\begin{figure*}[t]
\centerline{\includegraphics[width=0.8\textwidth]{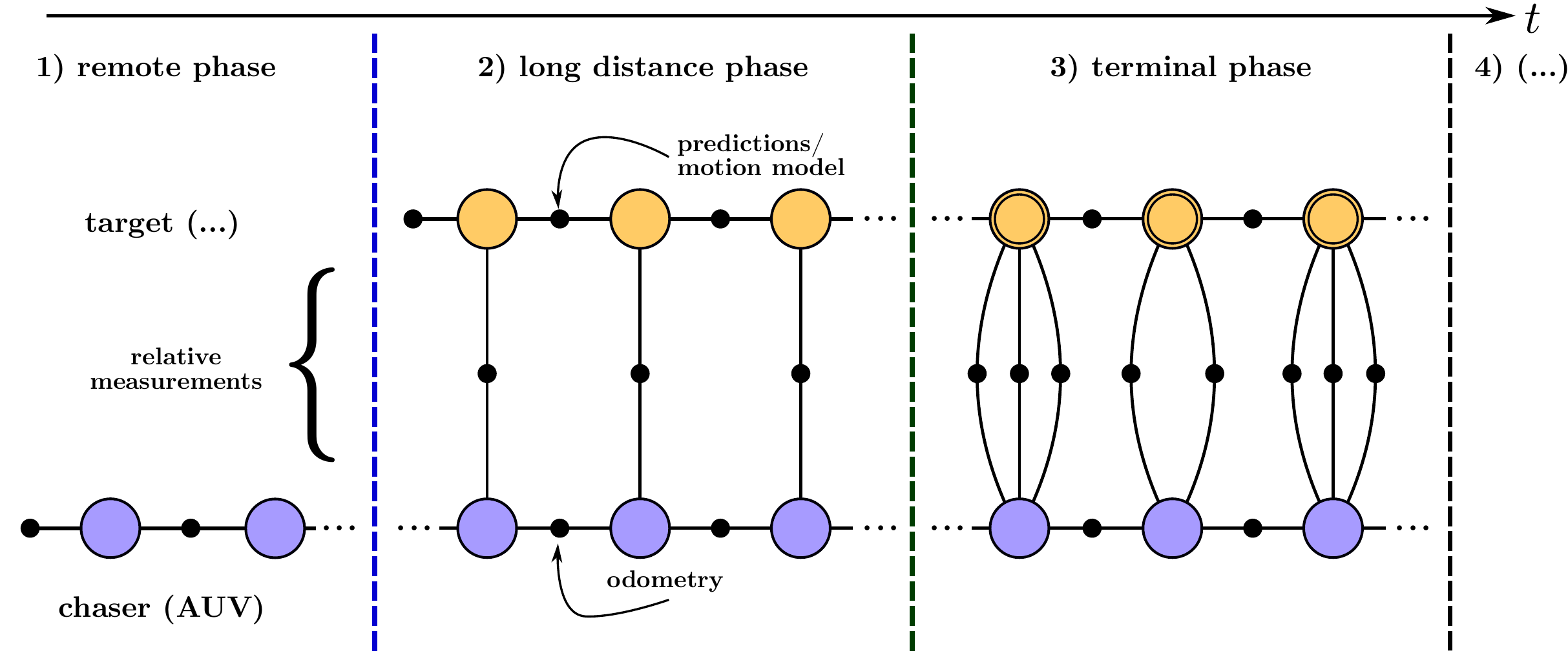}}
\caption{Factor graph representation of the STERN problem from a general proximity operation scenario. As the chaser navigates through the different phases of the prox-ops scenario to reach the target, the underlying state estimation problem will evolve with respect to the information available to constrain the relative state between agents.}
\label{fig:general_proxops_graph}
\end{figure*}

As mentioned before, regardless of the specific objective of any underwater docking mission, these all involve proximity operations between two agents: the chaser being the actively controlled agent, and the target acting as the objective. 
An illustration of a general underwater prox-ops scenario is shown in Figure \ref{fig:general_docking_phases}. The chaser is tasked with navigating towards and engaging a target (which varies depending on the prox-op at hand) by completing four phases that are determined by their specific sensing modalities and tasks to be executed. These are 1) the remote phase, 2) long-distance phase, 3) terminal phase, and 4) the joint phase.

\subsubsection{Remote phase}
As the prelude to the proximity operation, in the remote phase the chaser waits for a cue to begin the operation. Depending on the target and the operation itself, the remote phase has different conditions but the common denominator between them is the lack of communication and/or direct measurements to the target. Once the chaser is, for example, queried by a mothership or has entered the acoustic range to its docking station, the subsequent long distance phase can begin.

\subsubsection{Long distance phase}
After initial contact, the chaser can start planning and navigating to intercept the target. Due to the long range to the target (up to several kilometers), this phase is characterized by being able to measure the target's state only through acoustic signals~\cite{bellingham_autonomous_2016}. Typical long distance sensors provide information such as range, bearing, or a combination thereof~\cite{bellingham_autonomous_2016}. It is important to highlight that throughout this phase, the full navigation state of the target is not entirely resolvable, since only coarse information is available. A change of representation (e.g., from point landmark to full 3D pose) is possible once the agents get closer together and transition to the terminal phase, where the chaser's sensor suite capabilities are improved. 

\subsubsection{Terminal phase}
Either to home into a docking station or mothership, or to latch into an intervention panel or another AUV, the terminal phase requires the chaser to navigate in close vicinity to the target. Since acoustic measurements are limited in accuracy and sampling rate, in this phase the chaser's sensor suite is enhanced with additional capabilities (e.g., cameras, electromagnetic sensors, etc.) that provide higher resolution information with higher accuracy and at a higher rate. The caveat, however, is that these new sensors are sensitive to different oceanographic conditions that render their effective range much smaller than that of acoustic measurements---up to only a few tens of meters---which limits the terminal phase to be executed at only very short ranges to the target.

\subsubsection{Joint phase}
The joint phase, as the remote phase, depends completely on the type of proximity operation that is at play. However, the joint phase can be generally described as a combined phase where the chaser and the target interact and work in a collaborative or uncooperative manner. This phase might involve hard-docking into a station to recharge batteries and offload data, recovery by a mothership, station-keeping for wireless data transferring, inspection of the target, or even latching unto another AUV for collaborative tasks. 

Together, these four phases can be represented by means of a factor graph and used to describe a generic underwater docking scenario.

\subsection{Simultaneous Trajectory Estimation and Relative Navigation Problem Formulation}

\begin{table*}[t]
\caption{Related work on underwater docking.}
\label{tab:related_work}
\resizebox{1\textwidth}{!}{%
\begin{tabular}{@{}|l|l|l|l|l|l|l|l|l|l|@{}}
\toprule
\textbf{Refs.} &
  \multicolumn{1}{c|}{\begin{tabular}[c]{@{}c@{}}\cite{feezor_autonomous_2001},\cite{page_underwater_2021}, \cite{lin_docking_2022}\\ \cite{fletcher_lab_2017}, \cite{liu_detection_2019}, \cite{singh_integrated_1996}\\ \cite{sarda_launch_2019}, \cite{ferreira_homing_2015}, \cite{mcewen_docking_2008},\\ \cite{cowen_underwater_1997}\end{tabular}} &
  \multicolumn{1}{c|}{\begin{tabular}[c]{@{}c@{}}\cite{lin_underwater_2022}, \cite{vandavasi_concept_2018}, \cite{kondo_passive_2012},\\ \cite{li_auv_2015}, \cite{park_experiments_2009}, \cite{zhang_terminal_2019},\\ \cite{yan_dynamic_2020}\end{tabular}} &
  \multicolumn{1}{c|}{\begin{tabular}[c]{@{}c@{}}\cite{fan_auv_2019}, \cite{evans_autonomous_2003}, \cite{kimball_artemis_2018}, \\ \cite{matsuda_resident_2019}, \cite{maki_docking_2018}\end{tabular}} &
  \multicolumn{1}{c|}{\cite{vallicrosa_autonomous_2016}, \cite{hurtos_loon-dock_2017}, \cite{zhou_single_2018}} &
  \multicolumn{1}{c|}{\begin{tabular}[c]{@{}c@{}}\cite{palomeras_autonomous_2014}, \cite{brignone_fully_2007}, \cite{palmer_vision_2009}, \\ \cite{fallon_relocating_2013}\end{tabular}} &
  \multicolumn{1}{c|}{\cite{sans-muntadas_navigation_2015}, \cite{vaganay_homing_2000}, \cite{teo_robust_2012}} &
  \multicolumn{1}{c|}{\cite{keane_autonomous_2020}} &
  \multicolumn{1}{c|}{\cite{lin_improved_2021}, \cite{wang_visual_2021}} &
  \multicolumn{1}{c|}{\cite{ruan_factor_2020}} \\ \midrule
  
\textbf{\begin{tabular}[c]{@{}l@{}}Chaser\\ tasks\end{tabular}} &
  \begin{tabular}[c]{@{}l@{}}perception,\\ control\end{tabular} &
  \begin{tabular}[c]{@{}l@{}}perception,\\ control\end{tabular} &
  \begin{tabular}[c]{@{}l@{}}perception,\\ localization,\\ control\end{tabular} &
  \begin{tabular}[c]{@{}l@{}}perception,\\ localization\end{tabular} &
  \begin{tabular}[c]{@{}l@{}}perception,\\ SLAM\end{tabular} &
  localization &
  localization &
  \begin{tabular}[c]{@{}l@{}}loc. \cite{lin_improved_2021},\\SLAM,\\control\end{tabular} &
  SLAM \\
  
\textbf{\begin{tabular}[c]{@{}l@{}}Target\\type\end{tabular}} &
  \begin{tabular}[c]{@{}l@{}}passive\end{tabular} &
  \begin{tabular}[c]{@{}l@{}}passive\\(dynamic\\\cite{zhang_terminal_2019},\cite{yan_dynamic_2020})\end{tabular} &
  passive &
  \begin{tabular}[c]{@{}l@{}}passive\\(dynamic\\\cite{zhou_single_2018})\end{tabular} &
  passive &
  passive &
  passive &
  passive &
  \begin{tabular}[c]{@{}l@{}}dynamic\end{tabular} \\

\textbf{\begin{tabular}[c]{@{}l@{}}Chaser\\type\end{tabular}} &
  \begin{tabular}[c]{@{}l@{}}full homing\\and nav suite\end{tabular} &
  \begin{tabular}[c]{@{}l@{}}constr.\\ homing suite\end{tabular} &
  \begin{tabular}[c]{@{}l@{}}full homing\\and nav suite\end{tabular} &
  \begin{tabular}[c]{@{}l@{}}full homing\\and nav suite\end{tabular} &
  \begin{tabular}[c]{@{}l@{}}full homing\\and nav suite,\\ (constr. \cite{fallon_relocating_2013})\end{tabular} &
  \begin{tabular}[c]{@{}l@{}}constr.\end{tabular} &
  \begin{tabular}[c]{@{}l@{}}full homing\\and nav suite\end{tabular} &
  \begin{tabular}[c]{@{}l@{}}constr.\end{tabular} &
  \begin{tabular}[c]{@{}l@{}}constr.\end{tabular} \\
  
\textbf{Scene} &
  sea &
  \begin{tabular}[c]{@{}l@{}}tank\\(sim \cite{yan_dynamic_2020})\end{tabular} &
  \begin{tabular}[c]{@{}l@{}}tank, sea\end{tabular} &
  \begin{tabular}[c]{@{}l@{}}sim, sea\end{tabular} &
  \begin{tabular}[c]{@{}l@{}}tank, sim \cite{brignone_fully_2007},\\ (sea \cite{fallon_relocating_2013})\end{tabular} &
  simulation &
  sea/field &
  \begin{tabular}[c]{@{}l@{}}tank,\\ (sea \cite{wang_visual_2021})\end{tabular} &
  \begin{tabular}[c]{@{}l@{}}sim,\\tank \cite{ruan_factor_2020}\end{tabular} \\

\textbf{Methods} &
  \begin{tabular}[c]{@{}l@{}}tracking\end{tabular} &
  \begin{tabular}[c]{@{}l@{}}tracking\end{tabular} &
  \begin{tabular}[c]{@{}l@{}}filtering,\\terminal\\ tracking\end{tabular} &
  \begin{tabular}[c]{@{}l@{}}filtering,\\model-based\\ relative pose\\ estimation\end{tabular} &
  \begin{tabular}[c]{@{}l@{}}filtering,\\model-based\\ relative pose\\ estimation,\\ batch \\ opt. \cite{fallon_relocating_2013}\end{tabular} &
  filtering &
  trilateration &
  \begin{tabular}[c]{@{}l@{}}batch opt.\end{tabular} &
  \begin{tabular}[c]{@{}l@{}}batch opt.\end{tabular} \\

\textbf{\begin{tabular}[c]{@{}l@{}}Target\\priors\end{tabular}} &
  \begin{tabular}[c]{@{}l@{}}full/partial\\ prior state,\\ (no prior \cite{ferreira_homing_2015})\end{tabular} &
  \begin{tabular}[c]{@{}l@{}}full/partial\\ prior state\end{tabular} &
  \begin{tabular}[c]{@{}l@{}}full/partial\\ prior state\end{tabular} &
  \begin{tabular}[c]{@{}l@{}}full/partial\\ prior state\\ and model\end{tabular} &
  \begin{tabular}[c]{@{}l@{}}full/partial\\ prior state\\ and model\end{tabular} &
  \begin{tabular}[c]{@{}l@{}}full/partial\\ prior state\end{tabular} &
  \begin{tabular}[c]{@{}l@{}}known depth\end{tabular} &
  \begin{tabular}[c]{@{}l@{}}full prior\end{tabular} &
  \begin{tabular}[c]{@{}l@{}}constant\\dynamics\end{tabular} \\ \bottomrule
\end{tabular}%
}
\end{table*}

Herein we make use of graphical models to build a probabilistic representation of the a general proximity operation. As one of the standard solutions for modern SLAM problems~\cite{dellaert_factor_2017}, we leverage factor graphs---a bipartite graphical model representing a joint probability distribution~\cite{koller_probabilistic_2009}---to model the general prox-ops scenario description. Variable nodes represent states or quantities of interest that we want to estimate, while factor nodes typically encode sensor information and are used to jointly constrain multiple variable nodes. The full factor graph representation of this problem, which we have coined Simultaneous Trajectory Estimation and Relative Navigation (STERN), is shown in Figure~\ref{fig:general_proxops_graph}.

The scenario's factor graph is composed of two main parts: i) a chain of variable nodes that represents the state history of the chaser (almost always an AUV), and ii) a parallel chain of variable nodes that represents the state history of the target (type varies from operation to operation).

Each of these chains is separately propagated across time by some type of odometric factor. For the chaser's chain, since this tends to typically be an AUV, the factors between each state tend to be odometry measurements provided by the AUV's onboard Inertial Navigation System (INS). For the more general type of target (dynamically active), the factors between each state can take the form of an expected motion model used to predict the target's motion.

Also shown in Figure~\ref{fig:general_proxops_graph} are the soft boundaries for the prox-ops phases. During the remote phase, we only have information on the chaser's chain and we propagate it as it navigates. After crossing to the long distance phase it is necessary to start representing the state of the target. Thus, we create the target's chain and jointly constrain it to the chaser's chain via the relative measurements captured by the chaser's sensor suite. The specific type of factors used entirely depend on the available sensors.

As the two agents get much closer and enter the terminal phase, the chaser's sensor suite capabilities tend to greatly improve. This is reflected by the additional relative constraints between the two chains. On top of the extra constraints, a change of representation for the state of the target can also be warranted; while a point landmark can suffice for the variable nodes in the target's chain during the long distance phase, it might be necessary to estimate a full 6-degree-of-freedom (6-DOF) pose or a navigation state with velocities during the terminal phase before crossing to the final joint phase. 

In the next section, we apply this same factor graph context to represent the prox-ops scenarios of the identified related works from to the available literature.

\subsection{Related Work and Research Gap}

\begin{figure*}[t]
\centerline{\includegraphics[width=\textwidth]{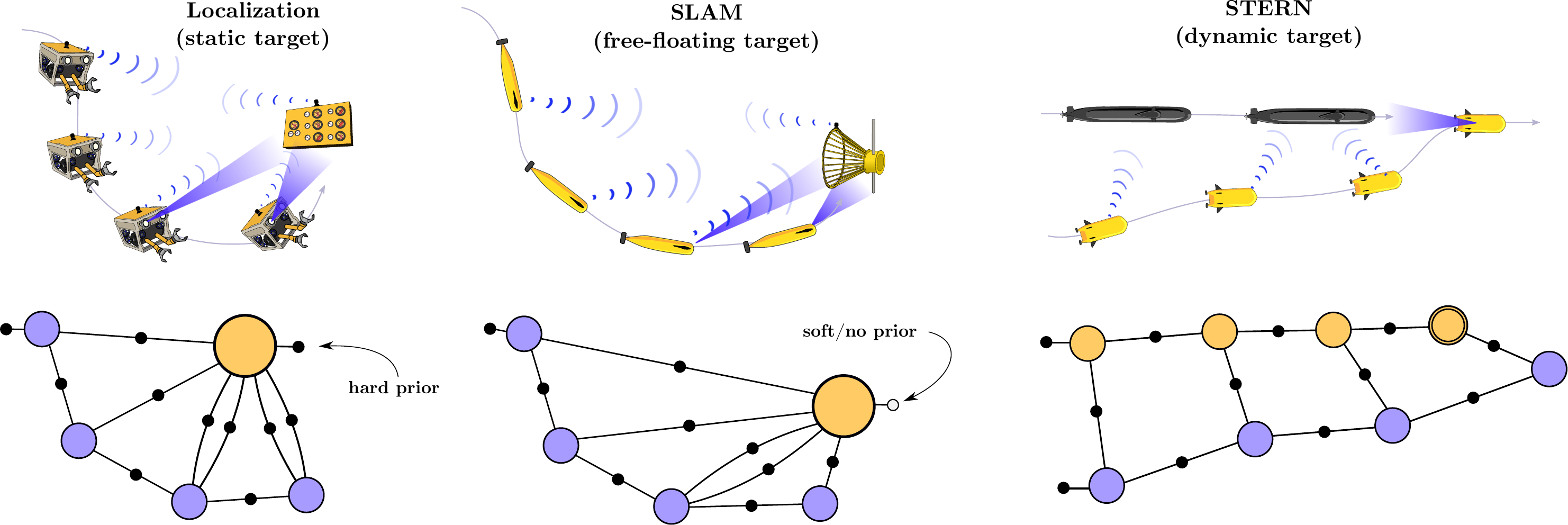}}
\caption{Depiction of the superset of different scenarios and chaser tasks identified in the literature (Table~\ref{tab:related_work}) together with their respective factor graph representation. Note that the closest scenario to our general version of a proximity operation (from Section~\ref{sec:background}) is clear to be that of solving the STERN task with a dynamic target.}
\label{fig:factor_graph_comparison}
\end{figure*}

To give the reader a full overview of the current state-of-the-art research on underwater proximity operations, which are mostly addressed as \textit{docking} or \textit{homing} missions, we compiled Table \ref{tab:related_work} with, to the best of our knowledge, all of what we considered to be work related to underwater robotic prox-ops. Table \ref{tab:related_work} is separated in columns that include works aiming at solving a similar scenario. The studied scenario for every column is described by the rows on the table which detail the following concepts: (1) the tasks executed automatically by the chaser during the prox-ops scenario that are described in the cited works; (2) the type of target, i.e., either passive or dynamically active; (3) the type of chaser---since in the case of underwater prox-ops the chaser is almost always an AUV, we make a distinction only with respect to their sensor suite---that is either a fully equipped AUV with a high-end navigation suite and a dedicated homing sensor suite (\textit{full homing and nav suite}), or a resource constrained AUV with, for example, a low-cost sensor suite and/or no dedicated homing suite (\textit{constr.}); (4) the environment or scene where methods were validated; (5) the methods used to tackle the different chaser tasks; and (6) important assumptions and constraints regarding the target, used to developed the methods.

Among the salient features that emerge from our surveyed table of related works is the popularity of tracking methods (first two columns) to solve passive (and dynamic in a couple of cases) target scenarios. Tracking, however, relies on directly coupling control with some sort of front-end perception task, without explicitly reasoning about the state of the target. In other words, the chaser navigates blindly towards the target by means of specific control cues without having an explicit state estimation module to provide high-level information regarding the target's position, attitude, or dynamics. Although sufficient for certain cases, tracking methods do not provide the necessary information for more complex and uncertain operations.

Among the methods that introduce an explicit perception and state estimation module (those using localization or SLAM), in contrast to the aforementioned tracking methods, a clear feature that we identify is that most references handle a passive target type, while only few handle the dynamically active type. We find this technical gap as the perfect candidate to showcase the use of the STERN framework detailed above. Although both target types are easily modeled using factors graphs, as briefly discussed in the previous section, we consider the dynamic target scenario a more general case than that of a passive one: a passive target, which in most cases was either a fixed or free-floating docking station or intervention panel, can be understood as a single variable node on the factor graph representing the state (which lives either $\in \R^{3}$ or $\in \SE(3)$, the Special Euclidean group in 3 dimensions) of the target; whereas a dynamically active target, since it evolves over time, will require the use of a variable node chain (as in Figure~\ref{fig:general_proxops_graph}) representing the state (in some navigation manifold $\in \calX$) tied together by factors computed with an adequate motion model. We depict in Figure~\ref{fig:factor_graph_comparison} how these scenarios and chaser tasks can be intuitively represented using factor graphs. This comparison emphasizes the generality of the prox-ops scenario with a dynamic target: those with a static or free-floating (passive) target can be solved by simply collapsing the target's state variable chain into a single variable node on the factor graph; whereas dynamic target scenarios land closer to the general prox-ops graph in Figure~\ref{fig:general_proxops_graph}.

As such, we choose herein to address the dynamically active target scenario during an underwater proximity operation and model it as a STERN problem. Although only a few authors have addressed this scenario (as seen from the table) its importance in providing the means to recover an AUV by a mothership while underway is nonetheless emphasized in all of them \cite{zhang_terminal_2019, yan_dynamic_2020, zhou_single_2018, hydroid_underwater_2012, ruan_factor_2020}, and whose concept has been thoroughly studied in \cite{watt_concept_2016}. 

In the present work, we land on a solution similar to \cite{ruan_factor_2020}, where the authors also use a factor graph-based approach to model and estimate the state of a dynamically active target. We borrow their same chaser AUV setup that uses Ultra-short Baseline (USBL) relative positions measurements, and a constant-dynamics assumption on the target's motion, and frame the scenario within the prox-ops context as a long-distance acoustic rendezvous with a moving mothership. In contrast to \cite{ruan_factor_2020} where they focus on the performance traits of batch optimization over filtering techniques, we focus on the modeling and assessment of the flexibility and modularity of factor graphs as a fundamental structure that can be used to model the scenario, and which can lead to the development of different flavored estimators---such as filtering, fixed-lag smoothing, or full incremental smoothing. We further build upon \cite{ruan_factor_2020} by providing a detailed roadmap on the steps required to deploy these solution on a real-world system, and provide thorough insight on its performance on both simulated, and field test data.

\section{Factor graphs as a flexible representation}
\label{sec:prop_approach}

\subsection{Background and Preliminaries}
Herein we define the notation and coordinate frame convention used throughout the paper for transformations on $\SE(3)$. We define the vector from point $a$ to point $b$ expressed in the coordinate frame $\Wf$ as $\Tran{\Wf}{b}{a} = -\Tran{\Wf}{a}{b}$. The rotation from the frame $\Wf$ to the frame A is written as $\Rot{\Wf}{\text{A}}= \Rot{\text{A}}{\Wf}^\top$, and to rotate the vector $\Tran{\Wf}{\text{P}}{\Wf}$ from frame $\Wf$ to the frame A we write $\Tran{\text{A}}{\text{P}}{\Wf} = {}^\text{A}_{\cancel{\Wf}}\Rot{}{}{}^{\cancel{\Wf}} \textbf{t}_{\text{P}/\Wf}$. We use this notation to define poses in $\SE(3)$ as rigid transformations in homogeneous coordinates. The pose of frame A expressed in frame $\Wf$ is written as
\begin{equation*}
    \Tfm{\Wf}{\text{A}} = \TfmMat{\Rot{\Wf}{\text{A}}}{\Tran{\Wf}{\text{A}}{\Wf}},
\end{equation*}
where $\bm{0}^\top \in \R^3$ is a vector of zeros; and the inverse pose of frame $\Wf$ expressed in frame A is computed as
\begin{equation*}
    \Tfm{\text{A}}{\Wf} = \Tfm{\Wf}{\text{A}}^{-1} = \TfmMat{\Rot{\Wf}{\text{A}}^\top}{-\Rot{\Wf}{\text{A}}^\top \Tran{\Wf}{\text{A}}{\Wf}}.
\end{equation*}
We express the velocity vector $\bm v$ of a frame A as seen from frame B, expressed in frame W as: $\Vel{W}{A}{B}$.

\begin{figure}[t]
\centerline{\includegraphics[width=0.6\linewidth]{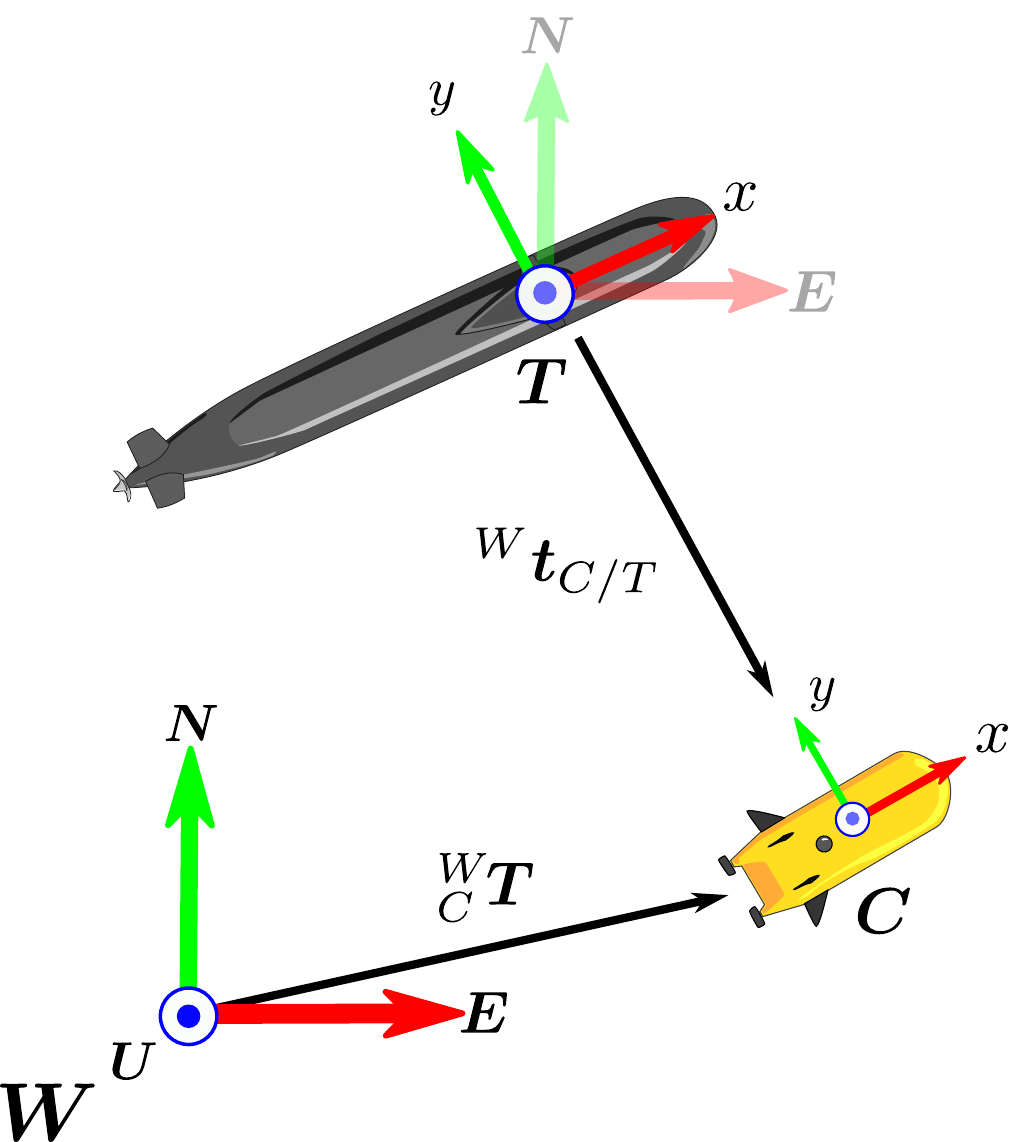}}
\caption{Reference frames involved in the USBL measurement procedure. Although the chaser does not initially have knowledge regarding the target's orientation, the use of the ENU frame as a common world frame $\Wf$ allows the chaser to rotate the relayed USBL measurement $\Tran{\Wf}{\Cf}{\Tf}$ from the world frame into its local frame C.}
\label{fig:usbl_corrdinate_frames}
\end{figure}

Throughout this work, three different coordinate frames (see Figure \ref{fig:usbl_corrdinate_frames}) are used: (1) the world frame $\Wf$ which is oriented with respect to the East North Up (ENU) convention, which is used as the main navigation frame for the chaser; (2) the chaser's local frame $\Cf$; and (3), the target's local frame $\Tf$.

\subsection{Problem Formulation}
Our objective in this section is to present the reader with the procedure of formulating the problem of estimating a set of latent variables given (1) a set of assumptions regarding the target, (2) a set of predictions generated based on a model of the target's motion, and (3) a set of measurements which provide observations on some or all the latent variables of interest---all within the factor graph framework.

\begin{figure*}[t]
\centerline{\includegraphics[width=\textwidth]{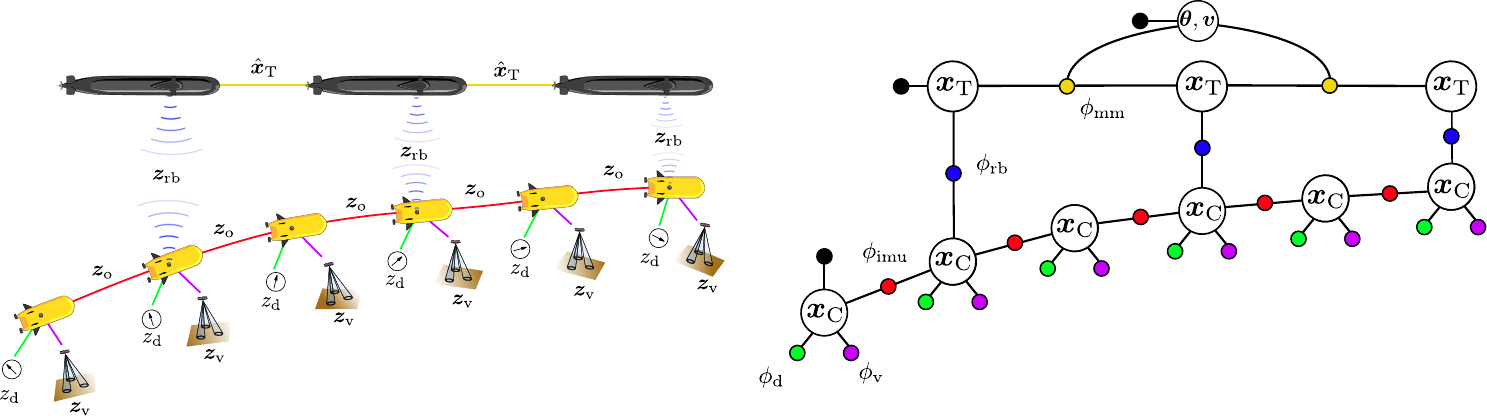}}
\caption{Side-by-side representation of the prox-ops scenario addressed in the present work. Left: pictorial description of the chasing AUV homing into the target mothership during the long distance phase, using a set of different measurements $\bm{z}_{(\cdot)}$ and predictions $\bm{\hat{x}}_{\text{T}}$; right: equivalent factor graph representation of the left figure, where different factors $\phi_{(\cdot)}$constrain the state of both the target $\tgtx$ and the chaser $\chx$.}
\label{fig:scenario_and_graph}
\end{figure*}

We study the prox-ops scenario of a resource constrained chaser AUV and a dynamically active target represented by a submerged mothership (submarine) during the long-distance phase where only acoustic measurements are available to measure the relative position between the agents, which is meant to simulate the starting phase of an AUV underway recovery mission~\cite{watt_concept_2016}. We tailor the measurements and assumptions involved in the scenario in such a way that it is possible for us to simulate and do field experiments with the resources at our disposal (e.g. in terms of platforms and sensors). Throughout the following sections, we provide a thorough description of how a basic state estimation solution for relative navigation for this scenario can be achieved using factor graphs. The scenario and its corresponding factor graph are shown side-by-side in Figure \ref{fig:scenario_and_graph}, highlighting the intuitive nature of factor graphs to model the scenario at hand.

The measurements involved in our scenario are explicitly depicted in Figure \ref{fig:scenario_and_graph} (left). As a resource constrained chaser, we assume the AUV uses a low-grade inertial navigation solution. This solution is composed of an attitude and heading reference system (AHRS)---measuring linear accelerations, orientation with respect to the world frame (ENU) and angular velocities---a Doppler velocity logger (DVL) for linear velocity measurements, and a pressure sensor to measure the depth of the AUV. For homing, the chaser's sensor suite only consists on an acoustic communication modem, through which it receives \textit{relayed} USBL acoustic measurements (detailed in Section \ref{sec:acoustic_message}) sent by the target after measuring them firsthand; thus, we assume the target mothership to be equipped with the necessary acoustic positioning and communications systems to measure a relative range and bearing to the AUV (i.e. USBL), and send it over to the AUV via the acoustic communication link.

Within our more specific problem formulation, the variable nodes for the chaser and the target in the factor graph shown in Figure~\ref{fig:scenario_and_graph} represent their state vectors, which take the following form:
\begin{align}
\label{eq:chaser_state}
    \chx &= [\Tfm{\Wf}{\Cf}\; \prescript{\Wf}{}{\bm v}_\Cf] \in \{\SE(3),\; \R^3\} = \calX, \text{ and } \\
    \tgtx &= \Tran{\Wf}{\Tf}{\Wf} \in \R^3,
\end{align}
where the target's state throughout the long distance phase is that of a 3D landmark. Since we are dealing with a dynamically active mothership, we adopt the notion that the vessel can keep a constant depth (as posed in \cite{watt_concept_2016}) to constrain the estimated landmark positions to a 2D plane.

We use a set of basic (yet warranted \cite{watt_concept_2016, hydroid_underwater_2012, ruan_factor_2020}) assumptions when modeling the dynamics of the target given that our first and foremost objective herein is neither solely maximizing estimation performance or model fidelity, but rather showcasing the suitability of STERN to easily accommodate generic sensor models and target motion models in a \emph{plug-and-play} manner. Firstly, we assume that it is highly likely to be dealing with a forward moving mothership during an AUV recovery process, due to the fact that a positive forward velocity will facilitate the subsequent prox-ops phases by granting higher control authority to both agents due to their hydrodynamic constraints~\cite{watt_concept_2016}; and secondly, we assume the scenario to take place in open waters, where no obstacles are present for the agents to avoid, and assume the target will keep a constant heading throughout (at least) the long distance phase, as argued by \cite{watt_concept_2016, hydroid_underwater_2012, ruan_factor_2020}. A straightforward way of including such constraints in the graph is by including variable nodes for estimating the target's unknown and constant velocity $v = ||\Vel{\Wf}{\Tf}{\Wf}|| = ||\prescript{\Wf}{}{}\bm v_{\Tf}|| \in \R$ and heading $\theta \in S^1$.

It is the interaction between the target's position $\tgtx$ and the two additional dynamics parameters $v, \theta$, together with some motion model, that allows us to explicitly reason about the motion of the target, essentially rendering the target to be an element of $\SE(2)$.

For this problem, the full set of the variables of interest to be estimated takes the form
\begin{equation}
\label{eq:map_estimation}
    \bm{\Theta} = \{ \mathbf{X}_\Cf, \mathbf{X}_\Tf, \theta, v\},
\end{equation}
where $\mathbf{X}_\Cf = \{{\chx}_i\}_{i=0:t_\text{f}}$ and $\mathbf{X}_\Tf = \{{\tgtx}_i\}_{i=0:t_\text{f}}$ are the entire state histories for the chaser and target, respectively, and $t_\text{f}$ denotes the final or most recent discrete timestep or keyframe. Thus, the objective of this work is to incrementally and in real-time compute the maximum a posteriori (MAP) estimate
\begin{equation}
\label{eq:map_argmax}
    \bm{\Theta}^* = \arg\max_{\bm{\Theta}} p(\bm{\Theta} | \mathbf{Z}),
\end{equation}
where our factor graph represents the joint probability distribution $p(\bm{\Theta} | \mathbf{Z})$ of our scenario at time $t_\text{f}$, and $\mathbf{Z}$ represents the set of all hitherto collected measurements. Note that this is the most general notation for the MAP estimator derived from a factor graph which results in fully smoothed estimates---since it optimizes the whole history of measurements and latent variables. However, by marginalizing away all but a fixed N amount of the latest variable nodes and measurements, i.e., a fixed window where $\mathbf{X}_\Cf = \{{\chx}_i\}_{i=t_{\text{f$-$N}}:t_\text{f}}$ and $\mathbf{X}_\Tf = \{{\tgtx}_i\}_{i=t_{\text{f$-$N}}:t_\text{f}}$ and $\bm Z_{k=t_{\text{f$-$N}}:t_\text{f}}$, we can effectively yield instead fixed-lag smoothed results; and naturally, by setting $\text{N}=1$, Equation~\eqref{eq:map_argmax} will act as a filter. Additionally, this can all be seamlessly done using the same back-end incremental solver that takes as input the same factor graph regardless of the method~\cite{kaess_isam2_2012, chiu_robust_2013, kaess_concurrent_2012}. This further supports our claims on the versatility of representing arbitrary prox-ops scenarios using factor graphs. 

Now that the foundation of the problem has been laid out, we do a deeper dive into the mechanics of the factor graph by describing the inner workings of each of our factors. 

\subsection{Building the factor graph}

Three different measurements are used to estimate the state of the chasing AUV $\chx$ with respect to a world frame: IMU, depth, and velocity measurements. 

Instead of directly relying on the AUV's navigation solution's output, we leverage the raw measurements coming directly from the navigation sensors in the AUV within the factor graph---a benefit of using a fully decoupled and general estimation framework. The output of the Inertial Measurement Unit (IMU) inside the chaser's AHRS is directly associated to the preintegrated IMU factor~\cite{carlone_eliminating_2014,forster_imu_2015} \tikzcircle[fill=red]{3pt}$\imuf$ and used to constrain the relative pose of the chaser $\Tfm{\Cf_i}{\Cf_{j}}$ between adjacent time steps ($t_i$ and $t_j$).
For absolute depth data, we compute a prior factor \tikzcircle[fill=green]{3pt}$\depthf$ on the depth of the chaser using pressure readings from the pressure sensor onboard the AUV. Similarly, a 3D velocity prior \tikzcircle[fill=dvl]{3pt}$\velf$ is computed using DVL measurements and added to the graph as a unary factor.

To constrain the relative position of the chaser to the target mothership, a USBL-based range and bearing factor \tikzcircle[fill=blue]{3pt}$\rbf$ calculated using the relayed USBL position measurements is used. Unrelated to any particular sensor or specific measurements, we use a motion model factor \tikzcircle[fill=yellow]{3pt}$\mmf$ that constrains adjacent target states by predicting the motion of the mothership using the additional variable nodes for $v$ and $\theta$.
Prior factors \tikzcircle[fill=black]{3pt}$\priorf$ are used on some variable nodes to encode the initial conditions for all the elements involved. 

In the following subsections, we present the theoretical details of each factor in Figure~\ref{fig:scenario_and_graph}.


\subsection{Acoustic Message as Range and Bearing Factor}
\label{sec:acoustic_message}

The USBL-based range and bearing factor \tikzcircle[fill=blue]{3pt}$\rbf$ is representative of the long distance phase of our scenario, since it encodes the only relative measurement the chaser obtains from its target until it crosses the hybrid sensing threshold. 

\begin{figure}[t]
\centerline{\includegraphics[width=3.5in]{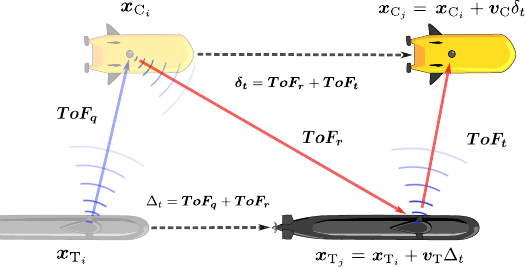}}
\caption{Pictorial description of the dynamic USBL measurement scenario. The arrows mark the path the acoustic pings travel during the measurement process. The target at $\tgtx{_i}$ sends a query ping to the chaser which arrives after $\text{ToF}_q$ (Time of Flight) seconds; the chaser at $\chx{_i}$ replies to the query and $\text{ToF}_r$ seconds pass before the ping returns to the target to complete one USBL measurement; the target computes the relative position fix $\Tran{\Wf}{\Cf}{\Tf}$ and relays it back to the chaser using the communication modem, which arrives at the chaser after having moved by a distance of $\bm{v}_{\text{C}}\delta_k$ from the position measured by the target.}
\label{fig:usbl_measurement}
\end{figure}

Before being added as a factor into the graph, the raw USBL measurement, or \textit{fix}, must endure a long journey in our scenario; as previously mentioned, we assume the use of \textit{relayed} USBL measurements. The acquisition procedure of the USBL measurement, depicted in Figure~\ref{fig:usbl_measurement}, works as follows: after querying the chaser by means of an acoustic ping, the target mothership receives a response ping from the chaser and computes a relative position using its onboard USBL acoustic positioning system; the same information as measured by the mothership is then relayed to the chaser via an acoustic communication link. 

The motivation behind this USBL setup is twofold. Firstly, even though USBL is commonly used for docking in an \textit{inverted} configuration (iUSBL), having only a transponder on the AUV and the transceiver as a top-side unit, is the most common USBL setup used for AUV navigation~\cite{leonard_autonomous_2016}, and it is precisely the only configuration we assume available for our resource constrained AUV. Secondly, covertness is oftentimes safety critical, so we avoid the scenario where the target mothership is actively pinging to prevent its location from being revealed to an arbitrary agent~\cite{watt_concept_2016}.

An issue to consider when dealing with relayed USBL measurements is the delay introduced into the system due to the transmission of the measurement during a dynamic scenario. As observed in Figure \ref{fig:usbl_measurement}, both agents actively move between each acoustic ping, meaning that some error is accumulated while the ping travels from one agent to the other. A brief analysis showed the induced error to be insubstantial (up to a few meters) with respect to the scale of our problem (100s to 1000s of meters). Thus, we leave the delay estimation and compensation as future work.

Since the fix is relayed to the chaser AUV exactly as it was measured by the mothership, a common coordinate frame must be used for the chaser to correctly rotate the received fix, resulting on a relative measurement akin to that of an iUSBL setup---where the chaser is in possession of the USBL transceiver and the target only has a transponder. Herein we use the ENU global reference frame (referred to as the world frame W) shown in Figure~\ref{fig:usbl_corrdinate_frames} as the common coordinate frame, and thus rely on the assumption that both chaser and target have a notion of this frame (e.g., by means of an onboard INS). Thus, the adjusted relayed USBL measurement takes the form
\begin{equation}
\label{eq:usbl_rotation}
    \Tran{\Cf}{\Tf}{\Cf} = - {\Tfm{\Wf}{\Cf}}^{-1} \cdot \Tran{\Wf}{\Cf}{\Tf},
\end{equation}
assuming that the target mothership has already pre-rotated the raw USBL measurement, originally measured with respect to the target's local frame as $\Tran{\Tf}{\Cf}{\Tf}$, into to the common reference frame W.

After obtaining the sought-after position fix $\Tran{\Cf}{\Tf}{\Cf}$ that represents the relative measurement of the position of the mothership with respect to the chaser, we can now assemble the range and bearing factor. 

The acoustic message ultimately encodes a range and bearing measurement $\z_{\text{rb}} = [z_{\text{r}}, \bm{z}_{\text{b}}]^\top \in \{\R^+, S^2\}$ representing the translation vector $\Tran{\Cf}{\Tf}{\Cf}$ between the chaser and target, where $S^2 = \{\bm{z} \in \R^3 : ||\bm{z}|| = 1\}$ denotes the 2-sphere manifold where the bearing measurements reside.

Using the chaser's $\chx$ and target's $\tgtx$ states, we use the range $h_{\text{r}}$ and bearing $\bm{h}_{\text{b}}$ measurement models to compute the predictions as
\begin{align}
\label{eq:range}
   h_{\text{r}} : \calX \times \R^3 \rightarrow \R^+&,   h_{\text{r}}(\chx, \tgtx) = ||\Tran{\Cf}{\Tf}{\Cf}|| = \hat{r}_{\Tf/\Cf},  \\[0.3em]
\label{eq:bearing}
    \bm{h}_{\text{b}} : \calX \times \R^3 \rightarrow S^2&, \bm{h}_{\text{b}}(\chx, \tgtx) = \frac{\Tran{\Cf}{\Tf}{\Cf}}{||\Tran{\Cf}{\Tf}{\Cf}||} = \Unit{\Cf}{\Tf}{\Cf}.
\end{align}

To build the range and bearing factor $\phi_{\text{rb}}$, we take the measurement prediction errors
\begin{align}
    \epsilon_{\text{r}} : \R^+ \times \R^+ \rightarrow \R&,  \epsilon_{\text{r}} = \hat{r}_{T/C} - z_{\text{r}}, \\
    \bm{\epsilon}_{\text{b}} : S^2 \times S^2 \rightarrow s^2&, \bm{\epsilon}_{\text{b}} = \bm{h}_{\text{b}}(\chx, \tgtx) \boxminus \z_{\text{b}},
\end{align}
and create the nonlinear Gaussian factor as
\begin{equation}
\label{eq:rbf}
\phi_{\text{rb}}(\chx, \tgtx) \propto \exp\left\{ -\frac{1}{2}
  \left| \left|
    \begin{bmatrix}
        \epsilon_{\text{r}}(\chx, \tgtx)\\
        \bm{\epsilon}_{\text{b}}(\chx, \tgtx)
    \end{bmatrix}
  \right| \right|^2_{\bm{\Sigma_{\text{rb}}}}
\right\},
\end{equation}
where $\boxminus : S^2 \times S^2 \rightarrow s^2$ is the on-manifold difference operator for bearing, 
$s^2 = \{\bm{b} \in \R^2 : ||\bm{b}|| < \pi\} \cup \{[\pi\, 0]^\top\}$ is the space that parameterizes $S^2$ without redundant degrees of freedom\cite{hertzberg_integrating_2013}, 
$\bm{\Sigma_{\text{rb}}}$ is the chosen covariance noise model, and $||(\cdot)||_{\bm{\Sigma}} = (\cdot)^\top \bm{\Sigma}^{-1} (\cdot)$ denotes the Mahalanobis distance.

Now that we have shown how we constrain the relative evolution of the state of the chaser and the position of the target using relayed USBL measurements, we continue into the dynamics associated to the motion of the target, and explain how we predict its motion.

\subsection{Constraining the dynamically active target}
\label{sec:mm_factor}
Herein we show how we leverage the prior knowledge we have regarding the dynamics of the target to predict its motion between timesteps---a task accomplished by the motion model factor \tikzcircle[fill=yellow]{3pt}$\mmf$. 

Based on the assumptions on the target's dynamics, we compute a motion model factor by associating consecutive target states $\tgtx{_i}$ and $\tgtx{_j}$, together with the most recent estimate of the constant heading $\theta$ and velocity $\bm v$. We compute a prediction of the target's subsequent state $\hat{\bm{x}}_{\text{T}}{_j}$ as
\begin{equation}\label{eq:mm_observation}
    h_{\text{mm}}(\tgtx{_i}, v, \theta) = \Rot{}{}(\theta) \tgtx{_i} + \bm{t}(v) = \hat{\bm{x}}_{\text{T}}{_j},
\end{equation}
where $h_{\text{mm}} : \R^3 \times \R \times S^1 \rightarrow \R^3$, $\Rot{}{}(\theta) \in SO(3)$ is the rotation matrix associated with the heading angle $\theta$, and $\bm{t}(v) \in \R^3$ is the translation due to the target's velocity $v$ over the time elapsed between states. The full expressions for $\Rot{}{}$ and $\bm{t}$ are shown in Appendix~\ref{sec:mm_jacobian}.

Using the prediction model from Equation~\eqref{eq:mm_observation}, we can compute the residual $\bm{\epsilon}_{\text{mm}}$ necessary to create the motion model factor $\mmf$. We define
\begin{equation}
\label{eq:mm_residual}
    \bm{\epsilon}_{\text{mm}} =  h_{\text{mm}}(\tgtx{_i}, \bm v, \theta) - \tgtx{_j},
\end{equation}
where $\bm{\epsilon}_{\text{mm}} : \R^3 \times \R^3 \times \R \times S^1 \rightarrow \R^3$. Using this residual, the motion model factor takes the form
\begin{equation}
\begin{aligned}
\label{eq:mm_factor}
\mmf&(\tgtx{_i}, \tgtx{_j}, \bm v, \theta) \\
    &\propto \exp\left\{ -\frac{1}{2}
  \left| \left|
    \bm{\epsilon}_{\text{mm}}(\tgtx{_i}, \tgtx{_j},\bm v, \theta)
  \right| \right|^2_{\bm{\Sigma_{\text{mm}}}}
\right\},
\end{aligned}
\end{equation}
with $\bm{\Sigma_{\text{mm}}}$ being the noise model for the factor. The Jacobian matrix associated to the motion model factor $\mmf$ and its derivation can also be found in Appendix~\ref{sec:mm_jacobian}.

\subsection{Chaser inertial navigation factors}
To estimate the chaser's state, we use three of the most common sensors found on inertial navigation solutions for AUVs \cite{leonard_autonomous_2016}: IMU, pressure sensor, and DVL. We avoid the use of the output of an INS and instead leverage the raw measurements as captured by the sensors themselves; this allows us to further generalize our approach by exempting the requirement of an onboard INS.

To constrain consecutive chaser states ${\chx}_i$ and ${\chx}_j$, we use the IMU's measured acceleration and angular rates to compute preintegrated IMU~\cite{carlone_eliminating_2014, forster_imu_2015} odometry measurements $\bm{z}_\Cf{_{ij}}$ and generate the factor
\begin{equation}
\begin{aligned}
&\imuf(\chx{_i}, \chx{_j}) \\
&\propto \exp\left\{ -\frac{1}{2}
  \left| \left|
    \left(\bm{\epsilon}_{\Delta\Rot{}{}_{ij}}^\top, \bm{\epsilon}_{\Delta\bm{t}_{ij}}^\top, \bm{\epsilon}_{\Delta\bm{v}_{ij}}^\top, \bm{\epsilon}_{\Delta\bm{b}_{ij}}^\top\right)^\top
  \right| \right|^2_{\bm{\Sigma_{\text{imu}}}}
\right\},
\end{aligned}
\end{equation}
where the $\bm{\epsilon}_{\Delta(\cdot)}$'s represent the residuals of the rotation $\Rot{}{}$, translation $\bm{t}$, velocity $\bm{v}$, and IMU biases $\bm{b}$.

The onboard pressure sensor measured the depth $z_{\text{d}}$ of the chaser. Thus, we use the observation model $h_{\text{d}}$ to access the estimated depth within the chaser state, which takes the form
\begin{equation}
     h_{\text{d}} : \calX \rightarrow \R,   h_{\text{d}}(\chx{_i}) ={}^{\Wf}t_z{_{\Cf/\Wf}}.
\end{equation}

The residual error $\bm{\epsilon}_{\text{d}}$ is computed similarly to the previous factors as
\begin{equation}
\label{eq:depth_error}
    \bm{\epsilon}_{\text{d}} : \mathcal{X} \rightarrow \R, \bm{\epsilon}_{\text{d}} =  h_{\text{d}}(\chx{_i}) \boxminus z_{\text{d}},
\end{equation}
and the prior depth factor $\depthf$ takes the form
\begin{equation}
    \depthf(\chx{_i}) \propto \exp\left\{ -\frac{1}{2}
  \left| \left|
    \bm{\epsilon}_{\text{d}}(\chx{_i})
  \right| \right|^2_{\bm{\sigma_{\text{d}}}}
\right\}.
\end{equation}
The factor's on-manifold subtraction $\boxminus$, as well as its associated Jacobian can be found in Appendix~\ref{sec:depthf_jacobian}.

Last but not least, we include the DVL's velocity measurements $\bm z_{\text{v}}$ as a prior velocity factor $\velf$ constraining only the chaser's velocity $\chx{_{\text{v}}}$ contained in its state. The factor takes the form
\begin{equation}
\velf(\chx{_i}) \propto \exp\left\{ -\frac{1}{2}
  \left| \left|
    \bm{\epsilon}_{\text{v}}(\chx{_i})
  \right| \right|^2_{\bm{\Sigma_{\text{v}}}}
\right\},
\end{equation}
where the measurement model $\bm{h}_{\text{v}}: \calX \rightarrow \R^3$ extracts the state's velocity components and is used to compute the residual $\bm{\epsilon}_{\text{v}} :\calX \rightarrow \R$ as
\begin{align}
     \bm{\epsilon}_{\text{v}} =  h_{\text{v}}(\chx{_i}) - \bm z_{\text{v}}.
\end{align}

In the following sections we provide insight on the implementation, verification, and experimental validation of our approach for both a simulated environment and with a real AUV platform.

\section{Implementation and Verification}
\label{sec:simulations}
The software implementation of the previously defined framework requires different modules to work together. We start this section by showing an overview of the developed STERN system, highlighting the different modules and the flow of information, and continue by presenting out the details concerning each module.

\begin{figure}[tb]
\centerline{\includegraphics[width=3.5in]{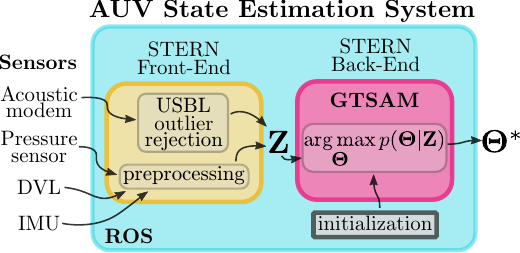}}
\caption{Block diagram showing the high-level module description and information flow for our STERN software implementation. From left to right, sensor measurements are processed by the both the front- and back-end modules to provide an optimized state estimate of our system $\bm{\Theta}^*$.}
\label{fig:high_level_diagram}
\end{figure}

\subsection{High-level Module Description and Information Flows}
Figure \ref{fig:high_level_diagram} shows a high-level view of the implemented STERN system. Our AUV state estimation system is developed using the Robot Operating System (ROS) middleware \cite{quigley_ros_2009} to interface the incoming sensor measurements with our STERN solution. The software is composed of two main blocks: the front-end and back-end.

The front-end is in charge of preprocessing the acquired sensor data to compute the set of measurements $\mathbf{Z}$ used by the back-end to create the factors presented in Section \ref{sec:prop_approach}. IMU, the pressure sensor's depth, and DVL, require little manipulation since the data output from the sensor is the measurement itself as used in the back-end module. However, the acoustic message (relayed USBL from the target) must be preprocessed to comply with the conditions required. USBL measurements are subject to different sources of noise that can cause spurious measurements, thus, an outlier rejection routine is added to avoid introducing such measurements into the factor graph.

The back-end module is responsible for building the factor graph using the set of preprocessed measurements $\mathbf{Z}$ and solving it to obtain an estimate of the variables of interest. Based mostly on functionality available on the Georgia Tech Smoothing and Mapping (GTSAM \cite{gtsam}) library---an open-source library with an efficient implementation of all the necessary algorithms for robotic perception using factor graphs---our back-end module solves Equation~\eqref{eq:map_estimation} and outputs the optimized set of system state $\bm{\Theta}^*$ that can be used by other modules further down the autonomy pipeline (e.g., planning, motion control, etc.).   

Below we provide specific details regarding each block in Figure \ref{fig:high_level_diagram}.

\subsection{STERN Front-end}
\label{sec:front_end}
Tasked with preprocessing the incoming sensor data to comply with the requirements of the back-end, the front-end must fulfill a few different procedures per sensor message before relaying the measurements to the back-end.

First of all, for each sensor, a ROS \textit{driver} is required to translate the incoming sensor data into a data type or message that can be manipulated. ROS drivers are typically only in charge of interpreting the sensor's communication protocol (e.g., serial, TCP/UDP) into a ROS-message, and broadcast it as a \textit{topic} inside the local ROS environment or \texttt{roscore} running on the AUV's computer. Once published inside the \texttt{roscore} any routine can subscribe to any relevant topic and use the sensor data it contains.

The incoming IMU data and the depth measured by the pressure sensor are both sent directly to the back-end after being processed by their respective ROS driver. The DVL velocities are measured by the sensor in the local frame ${}^\Cf \bm{v}_{\Cf/\Wf}$, however, since we are using them as a prior factor on the chaser's state (defined in Equation~\eqref{eq:chaser_state}), we rotate them into the world frame W as
\begin{equation}
    \bm{z}_v = {}^{\Wf} \bm{v}_{\Cf/\Wf} = \RotHat{\Wf}{\Cf} {}^\Cf\bm{v}_{\Cf/\Wf},
\end{equation}
where $\RotHat{\Wf}{\Cf}$ is the most recent orientation estimate of the chaser with respect to the world frame.

\begin{figure}[tb]
\centerline{\includegraphics[width=0.9\linewidth]{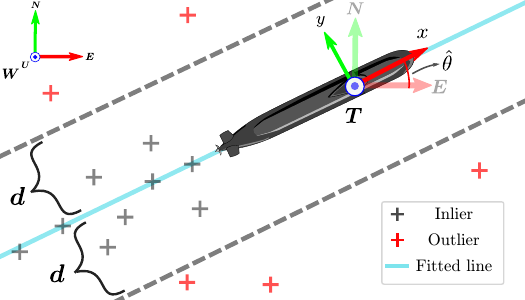}}
\caption{RANSAC-based outlier rejection example. New USBL measurements are checked for consistency with previously acquired measurements to avoid including spurious information in our back-end estimator.}
\label{fig:ransac_example}
\end{figure}

The USBL range and bearing measurement requires the most preprocessing among all of our sensors. We run an outlier rejection scheme based on the assumptions regarding the target's motion using Random Sampling Consensus (RANSAC) \cite{fischler_random_1981}. The outlier rejection routine, exemplified in Figure \ref{fig:ransac_example}, consists of calculating the Euclidean distance of the new measurement to a 2D line that is computed via RANSAC using a set of $M$ previously acquired fixes, and rejects it as an outlier if the distance exceeds a certain threshold $d$. The number of consecutive measurements $M$ can be modified to relax the constant heading constraint on the target's trajectory to add robustness against, for example, a slightly curved trajectory. After the evaluated fix is considered to be an inlier it is sent to the back-end.

\subsection{STERN Back-End}
Once the front-end has preprocessed all of the measurements to the right format, we can use them to compute our factors and build the factor graph. The factor graph is populated using a mix of predefined and custom factors in GTSAM. 

\subsubsection{Range and bearing factor}
\label{sec:range_bearing_factor} 
we leverage the built-in \texttt{BearingRangeFactor} available in GTSAM to compute the range and bearing factor $\phi_{\text{rb}}$ detailed in Equation~\eqref{eq:rbf}. In order to compute the factor, we calculate the range $z_{\text{r}}$ and bearing $\bm{z}_{\text{b}}$ as in Equation~\eqref{eq:range} and \eqref{eq:bearing}, respectively. We use a diagonal Gaussian noise model 
\begin{equation}
    \bm{\Sigma}_{\text{rb}} = \text{diag}(\sigma_{\psi} , \sigma_{\theta}, \sigma_{\text{r}}),
\end{equation}
where $\sigma_{\psi} = \sigma_{\theta} = 0.0175 \left[\frac{\text{rad}}{\text{m}}\right] \cdot z_{\text{r}}$ and $\sigma_{\text{r}} = 0.1 \left[\text{m}\right]$ are the standard deviations for the measured azimuth $\psi$, zenith $\theta$, and range $z_{\text{r}}$, respectively, of the spherical measurement.

\subsubsection{Preintegrated IMU factor} 
An efficient implementation of the IMU factor described in \cite{forster_imu_2015} can be found in the GTSAM library. We make use of the \texttt{CombinedPreintegratedMeasurements} to accumulate the incoming IMU acceleration and angular rates from the front-end module. These IMU measurements are preintegrated to compute an odometry estimate using the noise models
\begin{align}
    \bm{\Sigma}_{a} =&\ \sigma_a  I_3\ ;\ \bm{\Sigma}_{\omega} = \sigma_{\omega} I_3 \\
    \bm{\Sigma}_{\text{b}\omega} =&\ \sigma_{\text{b}\omega} I_3\ ;\ \bm{\Sigma}_{\text{b}a} = \sigma_{\text{b}a} I_3,
\end{align}
where 
\begin{align*}
\sigma_a &= 5.9898e^{-2}\left[\frac{m}{s^2}\frac{1}{\sqrt{Hz}}\right] \\
\sigma_{\omega} &= 1.0471e^{-5} \left[\frac{rad}{s}\frac{1}{\sqrt{Hz}}\right]
\end{align*} 
are the noise on the linear acceleration and angular rate, respectively, and
\begin{align*}
    \sigma_{\text{b}a} &= 5.0411e^{-2}\left[\frac{m}{s^3}\frac{1}{\sqrt{Hz}}\right] \\
    \sigma_{\text{b}\omega} &= 3.3936e^{-4}\left[\frac{rad}{s^2}\frac{1}{\sqrt{Hz}}\right]
\end{align*}
are the noise on the respective bias' random walk. The sigmas used for our IMU were computed as specified in \cite{rehder_extending_2016}. The computed odometry and uncertainty are included in the \texttt{CombinedImuFactor} to constrain consecutive chaser states, and compensate for IMU biases. 

\subsubsection{DVL velocity factor} prior factors constraining the most common variables in robot perception problems are readily available on GTSAM. We used the version of the \texttt{PriorFactor} which constrains a 3D velocity to add the DVL measurement into the factor graph. The covariance for the measured velocity $\bm{\Sigma}_{v}$ is provided by the sensor's driver in real time.

\subsubsection{Depth factor} since GTSAM has no predefined prior factor that constraints only one degree of freedom of a 6-DOF pose, one of the custom factors we designed was the \texttt{DepthFactor}. To develop a custom factor, we used a template for custom prior factors, where only the error function in Equation~\eqref{eq:depth_error} and the Jacobian of the observation model was specified (as detailed in the Appendix \ref{sec:depthf_jacobian}). We use the manufacturer-specified standard deviation $\sigma_d = 0.01 \left[\text{m}\right]$ for the pressure sensor as the noise model.

\subsubsection{Motion model factor} we created a custom factor to encode the constraints on the target's motion. This factor is the only one that is not associated with any measurement. Instead, the specified residual function takes the form of Equation~\eqref{eq:mm_residual}, which only relates variables inside the factor graph. The related Jacobian is implemented as derived in Appendix~\ref{sec:mm_jacobian}. The noise model for the factor $\bm{\Sigma}_{\text{mm}} = \text{diag}(\bm{\sigma}_{t}, \sigma_v, \sigma_{\theta})$, assigns the level of uncertainty on the estimated target's position $\bm{\sigma}_{t}$, velocity $\sigma_v$, and heading $\sigma_{\theta}$---where the specific values for the standard deviations are left as a tuning parameter that can be used to relax the constraints on the linear and constant motion assumptions. 

For every USBL measurement received by the back-end, a new \textit{keyframe} is computed. These keyframes consist of an instance of a \texttt{NonlinearFactorGraph} created using one of each of the aforementioned factors. Between keyframes, we bundle the IMU measurements by preintegrating them, and store the latest measured depth and velocity of the chaser until a new USBL fix is received.

Once the keyframe is generated, the factor graph is solved using a nonlinear optimization technique, which requires setting initial conditions for the variables of interest as a starting point for the iterative routine. For the chaser, we compose the previously optimized state estimate ${\chx}_i^*$ with the computed odometry $\z_{\Cf_{ij}}$ from the preintegrated IMU measurements as
\begin{equation}
  \hat{\bm{x}}_{\Cf_{j}} = {\chx}_i^* \oplus \z_{\Cf_{ij}},
\end{equation}
to calculate an initial value of its state. We use the time elapsed between keyframes $\Delta t$ to propagate the previous position of the target ${\tgtx}^*_{i}$ by means of the motion model in Equation~\eqref{eq:mm_observation}. Another possibility would be to use the measured range and bearing $\bm{z}_{\text{rb}{_j}}$ and the estimated chaser's state $\hat{\bm{x}}_\Cf{_j}$ to calculate the estimated target position, but we use the previous method for simplicity.

Last, after building the keyframe with all of our factors, and specifying the initial condition for the optimization, we call GTSAM's implementation of iSAM2 \cite{kaess_isam2_2012} to incrementally solve this problem using the Bayes tree---which provides efficient performance by reducing the number of computations associated with the inference process. The remaining two variables, the target's velocity $v$ and heading $\theta$, require no per-keyframe initial value since they are assumed constant.

\subsection{Initialization}
\label{sec:initialization}
The procedure detailed above is repeated for every keyframe. To initialize the factor graph, however, we must compute prior factors that constrain the initial states for the chaser's state $\bm{x}_{\Cf_0}$, the target's position $\bm{x}_{\Tf_0}$, as well as the target's velocity $v$ and heading $\theta$. 

The practical implementation of our solution is meant to be ran as a \textit{backseat} estimator that can run in parallel with the chaser's navigation filter. We leverage this setup and tap into the navigation filter's estimates to compute a prior factor on the first chaser state $\bm{x}_{\Cf_0}$ using the accumulated uncertainty as the initial noise $\bm{\Sigma}_{\Cf_0}$. We are able to use this measure of the initial uncertainty because we are mostly concerned about the relative pose estimation between agents. However, if we were to include Global Navigation Satellite System (GNSS) information as unary factors, we would be required to set this uncertainty to a significantly low value to anchor the initial position of the chaser on the global frame. 

The procedure to initialize the variables related to the target is more involved due to the lack of prior knowledge available. Thus, we use the assumptions on the linear trajectory and constant velocity to initialize the position $\bm{x}_{\Tf_0}$, velocity $v$ and heading $\theta$.

The initialization procedure follows the next steps: (1) we use the first $N \geq 2$ USBL range and bearing measurements $\bm{z}_{\text{rb}}$ to fit a 2D line as detailed in Section \ref{sec:front_end}; (2) we use this fitted line as a surrogate for the target's trajectory and use it to calculate an initial heading angle $\hat{\theta}$ as illustrated in Figure \ref{fig:ransac_example}; (3) we iterate through the set of $N$ measurements and calculate a velocity $\hat{v}_k$ for every two consecutive measurements, and average them to compute a prior velocity for the target $\bar{v} =\frac{1}{N-1}\sum_{k=0}^N \hat{v}_k$; (4) the last (i.e. the most recent) of the bundled $N$ measurements is used as a the prior target's position $\hat{\bm{x}}_{\Tf_0}$ and as the first keyframe to be optimized. Both $N$ and the initial noise models for the prior factors are a design parameter that can be tuned to provide the best results.

\section{Results and Discussion}
\label{sec:results}
For validation and demonstration of our implementation, we analyze the results of both simulated experiments and field experiments.

\subsection{Simulations}
The first set of validation tests were done on a simulated environment. To provide insight on the behavior and performance of our implementation, two model scenarios were designed with different conditions on the relative trajectories between agents, and both qualitative and quantitative results were obtained to understand the characteristics of our STERN solution.

\begin{figure}[t]
\centerline{\includegraphics[width=0.9\linewidth]{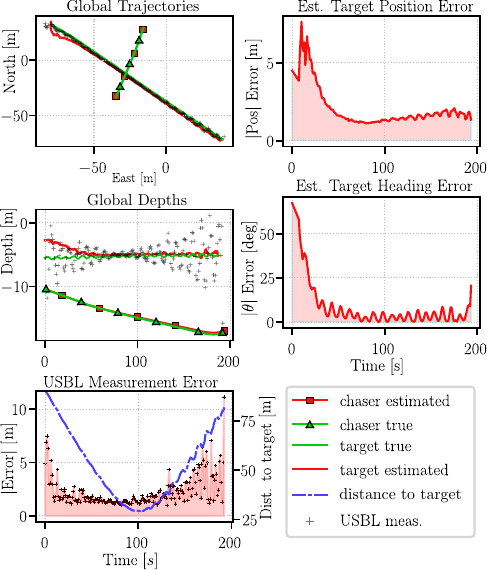}}
\caption{ Resulting 2D trajectories, depths, and absolute errors of the simulated perpendicular scenario.}
\label{fig:perpendicular_results}
\end{figure}

The first scenario, illustrated in Figure~\ref{fig:perpendicular_results} (top-left), consists on perpendicular trajectories between the agents: starting with a large relative distance, the agents travel on a straight line at a different depth, intersect at an arbitrary point in their trajectories, and continue on their way until reaching the end of the scenario.

The second scenario is shown in Figure~\ref{fig:parallel_results} (top-left). As opposed to the first scenario, here the agents travel along the same direction as to resemble parallel trajectories. The scenario begins with a large relative range between the agents, with the target far behind the chaser. The target then catches up to the chaser, continues on a parallel trajectory, and eventually overtakes the chaser---once again creating a big gap between agents.

Only USBL position measurements and the chaser's simulated navigation filter measurements were used for verification. The delay on the relayed USBL measurements was simulated by republishing the USBL fix measured by the target mothership after waiting $\delta_k$ seconds, i.e., the retransmission delay depicted in Figure \ref{fig:usbl_measurement}. The pinging rate was configured to a rate of $1$Hz. The noise on the dead-reckoned estimates for the chaser was set to a constant, low, known value to ensure that the error on the estimates was mostly caused by USBL measurement uncertainty---done so to facilitate the verification process.

As our simulated environment we used the Stonefish simulator~\cite{cieslak_stonefish_2019}---a marine robotics simulator compatible with ROS that allows for the simulation of the most common sensors used on underwater robots. Extensively used inside our research project~\cite{stenius_system_2022}\footnote{\href{https://github.com/smarc-project/smarc_stonefish_sims}{https://github.com/smarc-project/smarc\_stonefish\_sims.}}, we leverage the simulated model of our AUV LoLo as the chaser, and use a second simulated AUV as a surrogate for the target mothership in our scenario.

\subsubsection{Qualitative simulation results}
\begin{figure}[t]
\centerline{\includegraphics[width=0.9\linewidth]{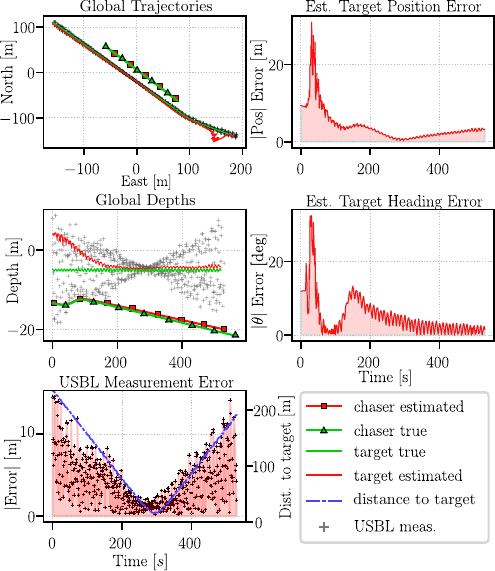}}
\caption{ Resulting 2D trajectories, depths, and absolute errors of the simulated parallel scenario. }
\label{fig:parallel_results}
\end{figure}

The results illustrated in Figures \ref{fig:perpendicular_results} and \ref{fig:parallel_results} show a comparison between the true and estimated poses obtained using our approach. It is important to underline that the estimate being shown is the per-keyframe, real-time latest estimate $\bm{x}_{t_i | 0:t_i}^*$ as computed by the back-end, and not the fully smoothed solution $\bm{x}_{t_i | 0:t_f}^*$, where $t_i \in [t_0, ..., t_f]$. This is the estimate that would be available to the robot during the real-time homing scenario with the current implementation of our approach.

Since we assumed almost no uncertainty on the chaser's navigation solution, we only evaluated the resulting estimates for the absolute target's position $\Tran{\Wf}{\Tf_i}{W}$ and orientation $\theta_i$ error expressed in the world frame for every $i$-th keyframe. 

Although both scenarios had different characteristics, the results for both missions show a similar behavior: after an initial divergence on the target's pose---spurred by purposefully using noisy priors---the optimized estimates successfully converge to the vicinity of the true pose as the number of measurements increases and the distance between the vehicles reduces.

\begin{figure*}[t!]
\centerline{\includegraphics[width=\textwidth]{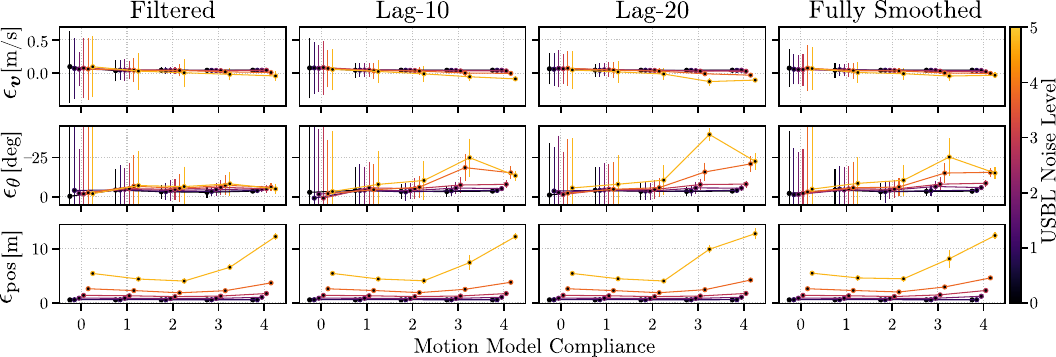}}
\caption{Plots showing the batch simulation results for the \textit{parallel} dataset. The error bars show the mean (black dots) and 3-$\sigma$ bounds (length of vertical line) for the computed velocity, heading, and positions errors ($\epsilon_{\bm v}, \epsilon_{\theta}, \epsilon_{\text{pos}}$, respectively). The x-axis of the plots corresponds to the degree of compliance with the motion model (Table~\ref{tab:compliance_levels}) and the colorbar to the right details the corresponding USBL noise level per plotted error bar (Table~\ref{tab:noise_level}).}
\label{fig:parallel_batch_results}
\end{figure*}

As noticed from the depth and USBL error plots in Figures \ref{fig:perpendicular_results} and \ref{fig:parallel_results}, there is a drop on the uncertainty of the USBL measurement as a function of the distance between the agents; as the distance between vehicles increases, so does the associated uncertainty of the fixes.
It is important to notice, however, that after converging, the estimate of the target's position is not significantly affected by the increasing uncertainty on the USBL fixes as the relative distance becomes large once more.

\begin{table}[b]
\centering
\caption{USBL noise levels used for batch simulations.}
\label{tab:noise_level}
\resizebox{\textwidth/2}{!}{%
\begin{tabular}{@{}c|ccc|ccc@{}}
\toprule
                                          & \multicolumn{3}{c|}{\textbf{USBL accuracy}}                                                    & \multicolumn{3}{c}{\textbf{Target initial uncertainty}}                                                  \\ \midrule
\multicolumn{1}{l|}{\textbf{Noise level}} & $\sigma_{\text{r}} [\text{m}]$ & $\sigma_{\theta} [\text{deg}]$ & $\sigma_{\psi} [\text{deg}]$ & $\sigma_{\text{pos}_0} [\text{m}]$ & $\sigma_{\bm{v}_0} [\text{m/s}]$ & $\sigma_{\theta_0} [\text{deg}]$ \\ \midrule
\textbf{\tikzcircle[fill=n0]{3pt} 0}       & 0.001                          & 0.001                          & 0.001                        & 0.1                                & 0.1                              & 0.1                              \\
\textbf{\tikzcircle[fill=n1]{3pt} 1}       & 0.01                           & 0.1                            & 0.1                          & 1.0                                & 0.2                              & 1.0                              \\
\textbf{\tikzcircle[fill=n2]{3pt} 2}       & 0.02                           & 0.3                            & 0.5                          & 2.0                                & 0.3                              & 3.0                              \\
\textbf{\tikzcircle[fill=n3]{3pt} 3}       & 0.5                            & 0.5                            & 1.0                          & 5.0                                & 0.5                              & 5.0                              \\
\textbf{\tikzcircle[fill=n4]{3pt} 4}       & 1.5                            & 1.0                            & 3.0                          & 10.0                               & 1.0                              & 10.0                             \\
\textbf{\tikzcircle[fill=n5]{3pt} 5}       & 5.0                            & 3.0                            & 5.0                          & 20.0                               & 1.5                              & 30.0                             \\ \bottomrule
\end{tabular}%
}
\end{table}

Another important observation regarding our simulated datasets is that, interestingly, a few model and control errors on the simulation environment caused the target AUV to have an oscillatory motion with respect to its heading and depth---violating our assumptions on the sea-keeping capabilities of the target---resulting in large oscillatory errors seen on the heading $\theta$ error plots in both figures. Nonetheless, we can still notice how the estimated heading, which is essential for our motion model, slowly approximates the true heading of the target. To further study the impact of the stringent assumptions on the target's motion embedded into the motion model factor, we present the following statistical results.

\subsubsection{Quantitative simulation results}
\label{sec:batch_simulations}
Using the datasets from the Stonefish simulator detailed above, we studied the sensitivity of our implementation to different USBL noise levels and the motion model's degree of compliance. 

Table~\ref{tab:noise_level} details the six different noise levels used for the USBL measurements, which were simulated by using the ground truth relative position of the target with respect to the chaser $\Tran{\Cf}{\Tf}{\Cf}$, converting it into a range and bearing measurement (Equations~\eqref{eq:bearing} and~\eqref{eq:range}), and corrupting each element with zero-mean Gaussian noise with its respective standard deviation from the table. Although the performance of acoustic measurements will always depend on the oceanographic conditions at the time of the measurements, we compile this table using numbers in agreement with our sensor's datasheet~\footnote{\href{https://www.evologics.com/product/s2c-r-7-17d-usbl-34}{EvoLogics S2C R 7/17 USBL datasheet.}} and other authors' findings~\cite{guerrero-font_usbl_2016}. Since we assume the target's state is initialized using USBL measurements, we increase the initial target's uncertainty accordingly. 

Table~\ref{tab:compliance_levels} presents the different levels of compliance given to the motion model. We define compliance as the tightness or rigidity of the constant dynamics constraints that is given to the predictions of the motion model factor. We manipulate this compliance by giving more or less weight to these assumptions through the noise model $\bm\Sigma_{\text{mm}}$ used to compute the squared Mahalanobis distance in Equation~\eqref{eq:mm_residual}.

\begin{table}[b]
\centering
\caption{Motion model uncertainties associated to the different levels of compliance.}
\label{tab:compliance_levels}
\resizebox{\textwidth/3}{!}{%
\begin{tabular}{@{}c|ccc@{}}
\toprule
                                                                    & \multicolumn{3}{c}{\textbf{Motion model uncertainty}}                                              \\ \midrule
\textbf{\begin{tabular}[c]{@{}c@{}}compliance\\ level\end{tabular}} & $\sigma_{\text{pos}} [\text{m}]$ & $\sigma_{\bm{v}} [\text{m/s}]$ & $\sigma_{\theta} [\text{deg}]$ \\ \midrule
\textbf{0}                                                          & 5.0                              & 1.0                            & 30.0                           \\
\textbf{1}                                                          & 2.5                              & 0.5                            & 10.0                           \\
\textbf{2}                                                          & 1.0                              & 0.25                           & 5.0                            \\
\textbf{3}                                                          & 0.5                              & 0.2                            & 2.5                            \\
\textbf{4}                                                          & 0.25                             & 0.15                           & 1.0                           
\end{tabular}%
}
\end{table}

Each of the USBL noise levels with each of the motion model compliance levels, yields a total of 30 different combinations that we use to benchmark the performance of our STERN solution. To further introduce flexibility on the target's dynamics, we use the same iSAM2 back-end solver to compute estimates using four different methods: (1) filtering, (2) smoothing (fixed-lag) with a window size of 10, (3) smoothing (fixed-lag) with a window size of 20, and (4) fully smoothing (real-time incremental solution). Filtering and fixed-lag smoothing forces the back-end to marginalize away the variable nodes constraining the motion model's constant velocity and heading ($\bm v, \theta$), whose initial conditions must be reinitialized with the best available information at the time. This renders piecewise constant dynamics constraints with the filtered solution, and only constant dynamics constraints within the N-keyframes window for the fixed-lag solutions. 

\begin{figure*}[t!]
\centerline{\includegraphics[width=\textwidth]{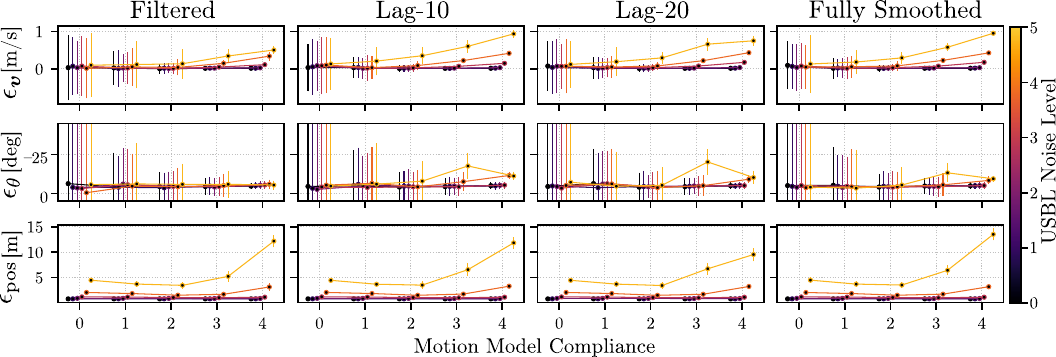}}
\caption{Plots showing the batch simulation results for the \textit{perpendicular} dataset. The error bars show the mean (black dots) and 3-$\sigma$ bounds (length of vertical line) for the computed velocity, heading, and positions errors ($\epsilon_{\bm v}, \epsilon_{\theta}, \epsilon_{\text{pos}}$, respectively). The x-axis of the plots corresponds to the degree of compliance with the motion model (Table~\ref{tab:compliance_levels}) and the color bar to the right details the corresponding USBL noise level per plotted error bar (Table~\ref{tab:noise_level}).}
\label{fig:perpendicular_batch_results}
\end{figure*}

Figures~\ref{fig:parallel_batch_results} and \ref{fig:perpendicular_batch_results} show error bars with the resulting errors in velocity $\epsilon_{\bm v}$, relative heading $\epsilon_{\theta}$, and relative position $\epsilon_{\text{pos}}$, obtained after running 100 iterations of each dataset per USBL noise level per degree of motion model compliance using each of the four estimation methods. Each colored line corresponds to a different USBL noise level (the colors follow those from Table~\ref{tab:noise_level}), the black circles and perpendicular colored lines detail the mean and the $3\sigma$ bound, respectively. The numerical results per dataset and method are reported in Tables~\ref{tab:perpendicular_batch_velocity_errors} to \ref{tab:parallel_batch_position_errors} in the Appendix; the bold numbers indicate the best results per USBL noise model within each method, and the highlighted show the best results per USBL noise model for all methods. By best we mean the most consistent results, which we calculate as the smallest mean error plus its respective standard deviation.

At first glance, the four methods exhibit a similar performance, where their mean accuracy is negatively impacted with increasing USBL noise and motion model tightness, while at the same time, counterintuitively, reducing the uncertainty of the results (smaller standard deviations). This suggests that the most consistent estimates are found by setting the degree of compliance somewhere on the mid- to tight-end of the range; this notion is supported by the fact that all the highlighted (best) results are achieved with compliance of 2 and above (from Tables~\ref{tab:perpendicular_batch_velocity_errors}-\ref{tab:parallel_batch_position_errors}). The results from both datasets show that the heading estimation benefits (although marginally) from the piecewise constant constraint introduced by the filtering method---which is able to track the oscillatory behavior of the target in the datasets more closely---whereas the results for the velocity suggest differently depending on the dataset. The position results point towards the existence of an optimal set of parameters somewhere in the middle as clearly seen from the banana-shaped curves in both Figure~\ref{fig:perpendicular_batch_results} and \ref{fig:parallel_batch_results}.

The key takeaways of this ablation study are: (1) although our assumptions on the motion of the target seem to be too constraining, our STERN solution can provide extra flexibility and achieve satisfactory performance by finding the degree of compliance that matches the scenario, and (2) the fully constant assumptions can be further loosened by solving the underlying graph using different optimization window sizes, without having a major impact on the performance of the estimates. 

The following section details the procedure and the results obtained from a real-world implementation of our method with our AUV demonstrator platform.

\subsection{Field Experiments}
\label{sec:experiments}

\begin{figure*}[t]
\centerline{\includegraphics[width=0.9\textwidth]{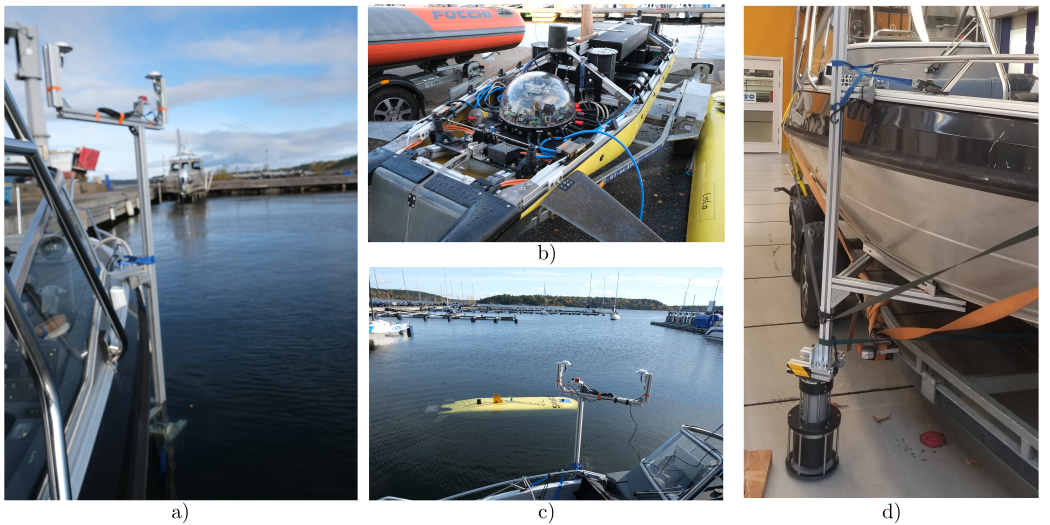}}
\caption{Images of the setup used for experimental verification: a) shows the side-deployment pole with the AHRS and dual-antenna GNSS at mounted on top, and the USBL top-side unit underneath; b) an image of the AUV LoLo before deployment, the USBL transponder can be seen above the glass dome housing; c) the service boat used as the target in our scenario next to the deployed AUV; d) simulation of the side-deployment pole with the mounted USBL top-side unit during its construction.}
\label{fig:experimental_setup}
\end{figure*}

After verifying the performance of our approach in the simulated environment, we further validate our approach by using datasets obtained from field experiments. In the following section, we give a detailed account of the experimental design, with all the relevant information regarding the hardware, additional software, and the deployments done with our vehicles.

As our chaser, we made use of our AUV LoLo presented in \cite{deutsch_design_2018}. The AUV LoLo (Figure \ref{fig:experimental_setup} b) and c)) is an experimental platform developed in-house for the Swedish Maritime Robotics Centre (SMaRC) \cite{stenius_smarc_2020}. As a prototype platform, the AUV LoLo is not fitted with a high-grade INS; instead, it relies on a relatively inexpensive navigation suite consisting of an SBG Ellipse 2 Micro AHRS, a Teledyne Pathfinder 600 DVL, a Keller Series 33x pressure sensor, and a GNSS module. These sensor measurements are fused using a particle filter running on a resource constrained Teensy 4.1 microcontroller. To communicate with the target mothership, our AUV is equipped with an Evologics S2C R7/17 USBL transponder, which doubles as an acoustic communication modem that is able to talk to the USBL \textit{top-side} unit mounted on the target mothership. The navigation sensor measurements and the received acoustic messages onboard the AUV are interfaced using ROS Noetic on an NVIDIA Jetson Xavier Orin running Ubuntu 20.04 (NVIDIA Linux 4 Tegra). LoLo is additionally equipped with 4G, WiFi, RF, and RC communication while on the surface for loading and running different missions while deployed.

As the target, we used the service boat (Figure \ref{fig:experimental_setup} a), d)) which serves as our mission control when deploying our different AUVs and surface vehicles. We fitted the service boat with a side deployment pole and mounted the USBL top-side transceiver unit below the waterline (Figure \ref{fig:experimental_setup} a) and d)). Rigidly attached to the USBL, at the top of the pole (see Figure \ref{fig:experimental_setup} a)), we attached a 3rd generation SBG Ellipse D AHRS with dual-antenna GNSS with true heading calculation to serve as our \textit{ground-truth} reference. Onboard the service boat, a Lenovo Thinkpad X1 Carbon (8th generation) laptop computer running ROS Melodic on Ubuntu 18.04 was used to interface the USBL unit and the AHRS. 

\begin{figure}[b!]
\centerline{\includegraphics[width=0.7\linewidth]{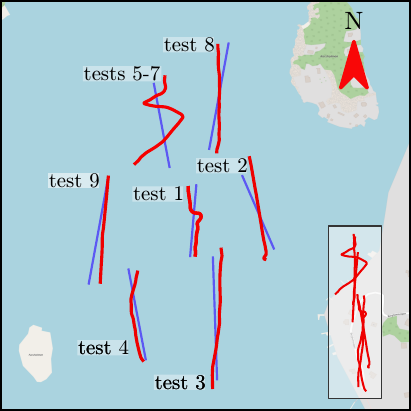}}
\caption{Exploded view of the labeled paths executed by the target (red) and approximated paths taken by the chaser (blue) during the nine experiments. The real arrangement of the different paths is shown on the lower right corner of the image.}
\label{fig:experiment_paths}
\end{figure}

The purpose of the field experiments was to validate our method of simultaneous trajectory estimation and relative navigation applied during the long distance phase of a prox-op scenario between a chaser AUV and a dynamically active mothership as the target. To do so, we deployed the our AUV in Baggensfjärden, a fjard east of Stockholm, Sweden. A total of seven deployments were done in one day of testing. Figure \ref{fig:experiment_paths} shows the trajectories followed by the service boat (the target in our scenario) and the (approximated) chaser, during the seven deployments, which were then processed into nine different datasets we used for offline validation.

\begin{figure*}[t]
\centerline{\includegraphics[width=\textwidth]{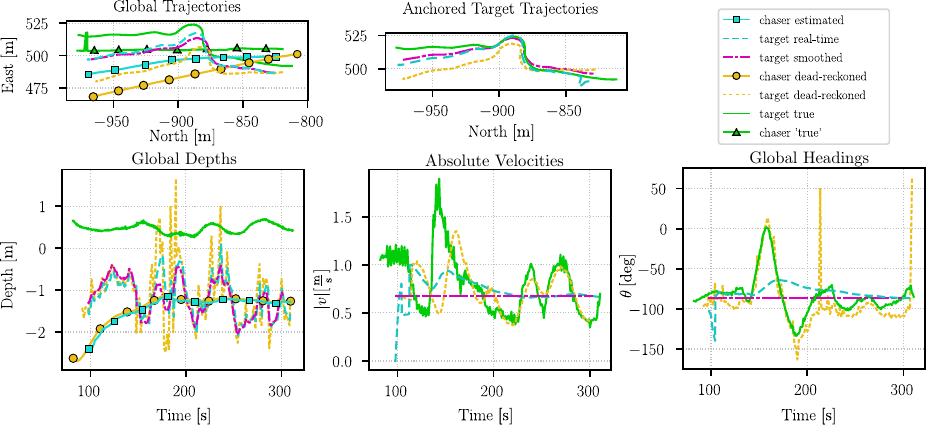}}
\caption{Plots showcasing the results obtained after processing the Test 1 dataset. The figure shows the global trajectories (true, real-time estimated, smoothed, and dead-reckoned) for both the target and the chaser, the true and estimated target's absolute velocity $|v|$ and heading $\theta$, and the anchored target trajectories used to quantify the position errors.}
\label{fig:test_1_results}
\end{figure*}

The AUV LoLo, as the chaser, was instructed to dive with a constant heading (from south-north and north-south only) and depth for an average of five minutes at a time during each of the seven deployments. The AUV was tasked to reply to every cue sent by the USBL topside unit. On the service boat, which represented the target mothership, we were tasked to: (1) simulate the behavior of a target whose motion model we have encoded in our factor graph (constant heading and velocity), and (2) to relay the measured USBL position fix to the AUV LoLo. Using a modified version of the ROS driver wrapping the Evologic's DMAC protocol\footnote{https://github.com/ocean-perception/dmac}, the onboard computer relayed every measurement instantly after reception. On response to any received message, the AUV would ping back to the service boat which would in turn trigger another position fix to be acquired by the topside unit. This measure-relay loop would continue automatically after an initial manually sent message, and we would timely stop it when the dive timeout on the AUV was a few seconds from expiring---this was done to avoid the AUV from pinging on air and damaging the transducer.

The scenario conditions for each of the deployments was meant to follow those used for simulations, however, a couple of reasons kept the service boat from following straight lines parallel or behind the AUV: (1) the lack of a visual interface for the USBL meant that we could only approximate the AUV's position underwater by reading the raw measurements, and (2) the limited controllability of the boat at low speeds ($\sim 4$ kts) made it hard to compensate the displacement caused by wind gusts. As consequence, in most of the collected datasets, the target behaves in an adversarial manner, voiding the assumptions of constant velocity and heading assumed in our method. 

The collected datasets were recorded using ROS' built-in logging system: \texttt{rosbag}. Two different \texttt{rosbags} were recorded during each deployment, one for each vehicle. On the AUV, the datasets recorded contained both the raw measurements from the navigation sensors, and the filtered dead-reckoning; additionally, the relayed USBL fix from the topside unit was converted into a ROS message and published inside the running \texttt{roscore} to be recorded together with the navigation messages. On the service boat's computer, a complementary \texttt{rosbag} was recorded with the raw and processed (by the USBL itself) USBL measurements together with the SBG's AHRS measurements containing the true heading, attitude angles, linear velocities, and GNSS measurements. Since both computers (LoLo's and the service boat's) had internet connection, we assumed their clocks to be synchronized to an acceptable extent. Each pair of per-deployment-\textit{rosbags} was merged offline into one single dataset that contained the timestamped data from both the service boat and the AUV LoLo.

Nine datasets were created from the seven deployments by splitting one of the deployments into three different sets (test 5-7 on Figure \ref{fig:experiment_paths}), done so by splitting each leg of the zig-zag maneuver to process separate diagonal scenarios.

The typical set of plots obtained from a single dataset is shown in Figure \ref{fig:test_1_results}. Three different estimates are compared in the plots to benchmark the performance of our method against the established ground truth (i.e. the target's track measured by its onboard GNSS). Due to the operation of the topside unit close to the surface of the water, and the shallow depth the AUV was diving on, the USBL's measurements of the AUV's depth were often spurious. Thus, in order to provide some sense of ground truth of the AUV's trajectory, we only use the XY-coordinates of the USBL position fix to provide an approximate measure of true chaser's position on the plane. This measure, however, is not sufficient to provide quantitative error estimates for the chaser's state. 

Figure \ref{fig:test_1_results} showcases the results obtained after processing the Test 1 dataset. Two estimates for the variables of interest are provided by our method: the real-time estimate $\bm{x}_{t_i | 0:t_i}^*$, computed per-keyframe by running the iSAM2 algorithm once, which relinearizes and updates the solution with the newly added factors; and the fully smoothed estimate $\bm{x}_{t_i | 0:t_f}^*$, which is calculated by a full back-substitution using the entire measurement history $t_i \in [t_0, ..., t_f]$. The dead-reckoned estimate of the target's state is computed by reprojecting the raw USBL measurements using the chaser's dead-reckoning provided by the onboard navigation filter; the velocity is then calculated using the distance and time traveled between two consecutive measurements and the heading is approximated by calculating the angle of the relative position vector between those same two measurements. The plots qualitatively compare all three estimates for both the target and the chaser with the ground truth for the target. 

\begin{figure}[b!]
\centerline{\includegraphics[width=\linewidth]{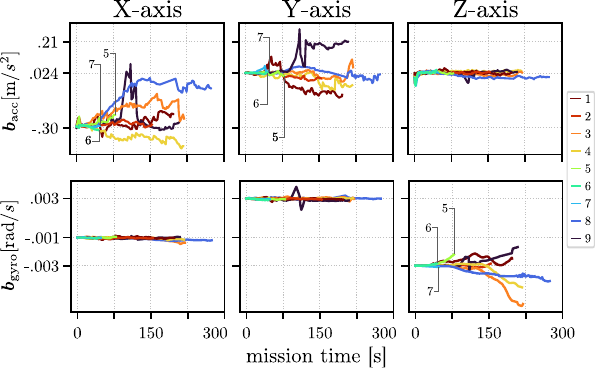}}
\caption{Time-series of the evolution of the estimated IMU accelerometer and gyroscope biases by our STERN back-end throughout the 9 different experimental deployments. The final values per test are detailed in Table~\ref{tab:imu_biases} in the Appendix}
\label{fig:imu_biases}
\end{figure}

\begin{figure*}[t!]
\centerline{\includegraphics[width=0.9\textwidth]{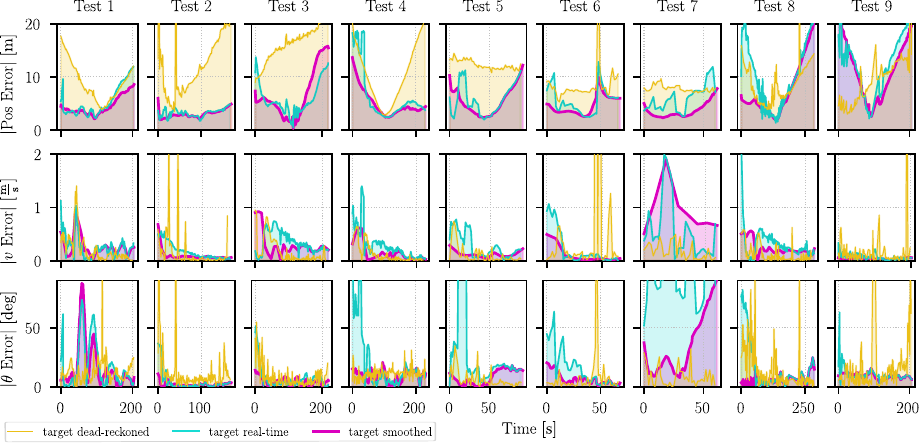}}
\caption{Plots for the absolute position, velocity $v$, and heading $\theta$ errors of the target per test. The plots compare the real-time estimates \tikzcircle[fill=slam]{3pt}, the smoothed estimates \tikzcircle[fill=smooth]{3pt}, and the dead-reckoned estimates \tikzcircle[fill=odom]{3pt}.}
\label{fig:error_plots}
\end{figure*}

\begin{figure}[b!]
\centerline{\includegraphics[width=\linewidth]{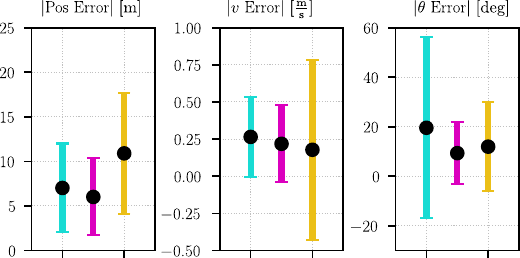}}
\caption{Error bars representing the mean (black circle) and standard deviation (bar length) of the sum of the position, velocity, and heading errors from the 9 tests quantified in Table \ref{tab:errors}. The figure depicts the total error statistics for the real-time estimates \tikzcircle[fill=slam]{3pt}, the smoothed estimates \tikzcircle[fill=smooth]{3pt}, and the dead-reckoned estimates \tikzcircle[fill=odom]{3pt}.}
\label{fig:error_bars}
\end{figure}

Before we continue with the analysis of the results it is important to highlight the performance of the dead-reckoned chaser estimates fused by the AUV's onboard filter: a substantial drift of the chaser's dead-reckoning---introduced by significant and unstable biases (Figure~\ref{fig:imu_biases})---can be appreciated by comparing the difference between the 'real' position of the chaser approximated by reprojecting the USBL measurements from the targets ground truth (in Figure \ref{fig:test_1_results} top-left corner). Since a great deal of caution had to be taken to avoid forcing the AUV to send acoustic pings on the surface, after the AUV started diving, we waited a short time before starting to ping the AUV from the service boat; unfortunately, this short time was enough for the dead-reckoning to drift and introduce an important offset in the initial estimates for the chaser's $\chx{_{_0}}$ and the target's $\tgtx{_{_0}}$ state. In order to give an equal fighting chance to the three estimates when computing the position errors, we decided to remove such offset by 'anchoring' the estimated target's track to the ground truth (top-middle plot in Figure \ref{fig:test_1_results}). We deem this an appropriate solution since our approach is mostly aimed at estimating the relative position between agents. 

From a visual analysis of the plots shown in Figure \ref{fig:test_1_results} we can derive a number of key insights. First, we highlight the fact that the estimated trajectory still follows the ground truth even when the underlying assumptions are violated, e.g., in the shown dataset the target is neither driving at a constant speed nor maintaining a constant heading; we attribute the resulting smooth, real target trajectory to the flexibility added to the constrains through the degree of motion model compliance (discussed in detail in Section~\ref{sec:batch_simulations}). Second, although we do not quantify it, it is possible to see how our method constrains the global drift of the dead-reckoning to some extent---a valuable characteristic to have regardless of our specific interest in the relative localization only. 

Hitherto we have discussed how a typical processed dataset looks like and pointed out some of the strengths of our method, we now shed some light on more of its advantages and limitations by means of a general error analysis computed using the information from all nine datasets. Figure \ref{fig:error_plots}, Table \ref{tab:errors}, and Figure \ref{fig:error_bars} complement one another to provide a general insight on the performance of our method. Table \ref{tab:errors} quantifies the mean and standard deviation of the absolute errors presented on Figure \ref{fig:error_plots}, and highlights the best performing result per variable and dataset. As expected, the smoothed estimates outperform the real-time values, however, there are a few instances where the dead-reckoned results outperform both of our method's estimated values for the target's velocity $v$ and heading $\theta$; this is explained by the fact that the target's motion model assumes constant velocity and heading, which, as explained before, is not the case for most of our datasets. This behavior is clear in Figure \ref{fig:test_1_results}: the reprojected raw USBL measurements on the dead-reckoned chaser's trajectory can track this unexpected behavior closer than the smoothed, constant estimates can. Another common observed limitation is the failed initialization of the heading angle for the target: the poor initial performance of the real-time estimates can be seen (for most tests) in Figure \ref{fig:error_plots}, influencing the general uncertainty of the real-time estimated heading quantified in Table \ref{tab:errors}. We observed that, typically, the first few USBL measurements tend to be noisier than average and, given that we use the first $N$ USBL measurements for initialization of both the velocity and the heading (Section \ref{sec:initialization}), it resulted in a failed initialization. This error is clearly compensated by the full back-substitution of the smoothed estimates. 

\begin{table}[t!]
\caption{Mean and standard deviation of the errors in Figure \ref{fig:error_plots}.}
\label{tab:errors}
\resizebox{0.45\textwidth}{!}{%
\begin{tabular}{@{}cc|c|c|c@{}}
\toprule
                   &      & \begin{tabular}[c]{@{}c@{}}Real-time\\ Estimated\end{tabular} & \multicolumn{1}{c|}{Smoothed} & \multicolumn{1}{c}{Dead-reckoned} \\ \midrule
\multirow{3}{*}{1} & pos & 5.34 $\pm$ 2.77 &\textbf{ 4.51 $\bm\pm$ 1.7} & 8.98 $\pm$ 3.53 \\
& $v$ & \textbf{0.26 $\pm$ 0.21} & \textbf{0.27 $\pm$ 0.2} & 0.22 $\pm$ 0.26 \\
& $\theta$ & 19.0 $\pm$ 20.09 & 16.81 $\pm$ 20.32 & \textbf{13.38 $\bm\pm$ 13.09} \\ \midrule
\multirow{3}{*}{2} & pos & 3.25 $\pm$ 0.73 & \textbf{3.04 $\bm\pm$ 0.72} & 12.71 $\pm$ 9.34 \\
& $v$ & 0.17 $\pm$ 0.14 & \textbf{0.1 $\bm\pm$ 0.14 }& 0.14 $\pm$ 0.36 \\
& $\theta$ & 3.68 $\pm$ 7.61 &\textbf{ 2.0 $\bm\pm$ 1.78} & 12.39 $\pm$ 22.63 \\ \midrule
\multirow{3}{*}{3} & pos & \textbf{5.77 $\bm\pm$ 2.89} & 6.39 $\pm$ 4.42 & 16.26 $\pm$ 3.2 \\
& $v$ & 0.35 $\pm$ 0.19 & 0.3 $\pm$ 0.24 &\textbf{ 0.16 $\bm\pm$ 0.23 }\\
& $\theta$ & 7.13 $\pm$ 9.03 &\textbf{ 4.29 $\bm\pm$ 3.35} & 8.05 $\pm$ 9.39 \\ \midrule
\multirow{3}{*}{4} & pos & 6.15 $\pm$ 4.82 & \textbf{4.91 $\bm\pm$ 2.46 }& 10.36 $\pm$ 5.48 \\
& $v$ & 0.28 $\pm$ 0.28 & 0.15 $\pm$ 0.18 & \textbf{0.12 $\bm\pm$ 0.16 }\\
& $\theta$ & 18.21 $\pm$ 23.22 & \textbf{9.5 $\bm\pm$ 4.84 }& 11.42 $\pm$ 4.9 \\ \midrule
\multirow{3}{*}{5} & pos & 5.72 $\pm$ 2.84 & \textbf{5.34 $\bm\pm$ 2.59} & 12.23 $\pm$ 0.81 \\
& $v$ & 0.25 $\pm$ 0.22 & \textbf{0.13 $\bm\pm$ 0.07} & 0.1 $\pm$ 0.16 \\
& $\theta$ & 43.25 $\pm$ 85.32 & 8.46 $\pm$ 6.14 & \textbf{6.03 $\bm\pm$ 6.98} \\ \midrule
\multirow{3}{*}{6} & pos & 5.51 $\pm$ 2.57 & \textbf{4.67 $\bm\pm$ 1.8} & 8.46 $\pm$ 4.13 \\
& $v$ & 0.23 $\pm$ 0.33 & \textbf{0.13 $\bm\pm$ 0.17} & 0.71 $\pm$ 2.43 \\
& $\theta$ & 15.88 $\pm$ 15.94 & \textbf{6.66 $\bm\pm$ 5.25} & 9.2 $\pm$ 25.55 \\ \midrule
\multirow{3}{*}{7} & pos & 6.16 $\pm$ 2.49 & \textbf{3.74 $\bm\pm$ 1.48} & 7.83 $\pm$ 0.65 \\
& $v$ & 0.56 $\pm$ 0.54 & 0.94 $\pm$ 0.41 & \textbf{0.18 $\bm\pm$ 0.15} \\
& $\theta$ & 119.07 $\pm$ 52.28 & 36.22 $\pm$ 26.11 & \textbf{8.99 $\bm\pm$ 8.57} \\ \midrule
\multirow{3}{*}{8} & pos & 10.15 $\pm$ 6.22 & \textbf{7.26 $\bm\pm$ 5.32 }& 8.82 $\pm$ 4.28 \\
& $v$ & 0.34 $\pm$ 0.3 & 0.25 $\pm$ 0.15 &\textbf{ 0.13 $\bm\pm$ 0.17} \\
& $\theta$ & 16.39 $\pm$ 20.75 & \textbf{6.13 $\bm\pm$ 5.08} & 16.07 $\pm$ 25.09 \\ \midrule
\multirow{3}{*}{9} & pos & 12.58 $\pm$ 5.36 & \textbf{11.63 $\bm\pm$ 5.11} & 9.96 $\pm$ 11.99 \\
& $v$ & 0.07 $\pm$ 0.06 & \textbf{0.03 $\bm\pm$ 0.02} & 0.15 $\pm$ 0.46 \\
& $\theta$ & 10.39 $\pm$ 6.52 & \textbf{9.26 $\bm\pm$ 2.15} & 14.49 $\pm$ 24.41 \\ \bottomrule
\end{tabular}%
}
\end{table}

Further challenges and errors are introduced due to unmodeled USBL uncertainties. Harsh outliers are still being observed and introduced to the factor graph (see Test 6 in Figure \ref{fig:error_plots} around the 50 second mark), sneaking past the RANSAC detection scheme (Figure \ref{fig:ransac_example}) which is only working on the 2D plane; instead, the procedure must actually occur in 3D space, i.e., the distance parameter $d$ has to describe the radius of a 3D cylinder instead of a 2D strip. Real USBL measurements also presented a more challenging noise behavior that our naive noise model (Section \ref{sec:range_bearing_factor}) does not take into account---a proper noise model is paramount for adequate sensor fusion.

In spite of all these limitations, however, our STERN approach shows promise by providing consistent estimates of the target's position, velocity, and heading---although the real-time estimates can waver during initialization---as understood from Figure \ref{fig:error_bars}. It is important to note that the dead-reckoned estimates for heading and velocity prove to be more accurate in some cases (see in Table~\ref{tab:errors}), caused by the fact the the motion model assumptions are violated in our estimator; this presents itself as a potential improvement on the estimated motion model variables, by appending this information as additional prior factors. However, in general, the dead-reckoned values are less consistent as seen by the larger standard deviations in Figure~\ref{fig:error_bars}) introduced by outliers and noisy measurements (error peaks in Figure \ref{fig:error_plots}); The significant standard deviation on the real-time estimates of the target's heading help us understand the shortcomings of our initialization routine and how we can improve our real-time values.

\subsection{Timing breakdown}
Some concerns could arise regarding the fact that a full smoothing solution, where the whole history of both agent's states, together with the whole history of sensor measurements and motion predictions, could significantly impact the computational complexity of the estimation process and render it impossible to work as an online solution on a real AUV. To answer these questions we provide a full timing breakdown in Table~\ref{tab:timing_breakdown} of the processes involved in the whole pipeline---from sensor measurement to an updated state estimate. 

\begin{table}[t]
\centering
\caption{Keyframe timing breakdown for the front- and back-end processes.}
\label{tab:timing_breakdown}
\resizebox{\textwidth/2}{!}{%
\begin{tabular}{@{}lcc@{}}
\toprule
\multicolumn{3}{c}{\textbf{Front-End}}                                                                                        \\ \midrule
\multicolumn{1}{l|}{\textbf{Process}}   & \multicolumn{1}{c|}{Average time [ms]} & \% of total                                \\ \midrule
\multicolumn{1}{l|}{IMU preintegration} & \multicolumn{1}{c|}{0.074$\pm$0.032}   & 0.85                                       \\
\multicolumn{1}{l|}{Depth processing}   & \multicolumn{1}{c|}{0.001$\pm$0.001}   & 0.02                                       \\
\multicolumn{1}{l|}{DVL processing}     & \multicolumn{1}{c|}{0.012$\pm$0.005}   & 0.14                                       \\
\multicolumn{1}{l|}{USBL front-end}     & \multicolumn{1}{c|}{0.18$\pm$0.053}    & 2.07                                       \\ \midrule
\multicolumn{3}{c}{\textbf{Back-End}}                                                                                         \\ \midrule
\multicolumn{1}{l|}{\textbf{Process}}   & \multicolumn{1}{c|}{Average Time [ms]} & \% of total                                \\ \midrule
\multicolumn{1}{l|}{Keyframe factors}   & \multicolumn{1}{c|}{0.195$\pm$0.038}   & 2.24                                       \\
\multicolumn{1}{l|}{iSAM2 graph solve}  & \multicolumn{1}{c|}{6.484$\pm$13.68}   & 74.31                                      \\
\multicolumn{1}{l|}{Update estimates}   & \multicolumn{1}{c|}{1.778$\pm$1.204}   & 20.37                                      \\ \midrule
                                        & \multicolumn{1}{l}{}                   & \multicolumn{1}{l}{}                       \\ \midrule
\multicolumn{2}{r|}{\textbf{Total time (average)}}                               & \multicolumn{1}{r}{8.724$\pm$14.12 [ms]}   \\
\multicolumn{2}{r|}{\textbf{Total time (upper bound)}}                           & \multicolumn{1}{r}{\textbf{53.726 [ms]}}   \\
\multicolumn{2}{l|}{\textbf{Total budget @ keyframing rate ($\sim$0.5Hz)}}            & \multicolumn{1}{r}{2000 [ms]}              \\
\multicolumn{2}{r|}{\textbf{Compute headroom (budget - total)}}                  & \multicolumn{1}{r}{\textbf{1946.238 [ms]}} \\ \bottomrule
\end{tabular}%
}
\end{table}

Table~\ref{tab:timing_breakdown} shows the mean and standard deviation of the time it took to run each process while processing all nine experimental datasets that are presented in the following Section~\ref{sec:experiments} on a laptop computer rocking an Intel i7-10510U CPU. With the USBL positioning system running at an average rate of 0.5Hz, the table shows the significant compute headroom available during each keyframe after the whole pipeline has ran. Naturally, if the solver were to be ran using only a subset of the history of measurements and states (e.g. filtering or fixed-lag smoothing), the iSAM2 graph solve step would consume less resources.

\section{Conclusions}
\label{sec:conclusions}
Autonomous underwater proximity operations are of paramount importance when working towards enhancing both the endurance and the capabilities of AUVs. Underwater prox-ops will enable AUVs to tackle target-based missions with arbitrary targets, a critical role in several modern academic and commercial applications.

In the present work, we took a holistic look at the problem of robotic underwater relative navigation and borrowed concepts used for spacecraft docking to frame it in a proximity operation framework. Proximity operations (or prox-ops) are specific missions that involve two different agents: a chaser and a target. The chaser must undergo a series of phases that are given by the specific sensor modalities available during its approach to the target. We specifically look at the problem of simultaneously estimating the trajectory of the target and its relative state with respect to the chaser (a problem we coined STERN in Section~\ref{sec:background}) during the different phases of a prox-ops scenario, and leverage the generality of factor graphs for state estimation to model any prox-ops scenario regardless of the type of target and chaser. 

After an extensive literature review, we categorized the current state-of-the-art on underwater proximity operations and framed the different scenarios applying a general framework based on modeling the information fusion problem using factor graphs to show how any arbitrary operation can be modeled with the same methodology; this is done to enable the straightforward comparison and reproduction of research within the same subject, and to serve as a foundation for incremental research. The literature review shed light on an existing technical gap in the field, namely the problem of dealing with dynamic targets, which we pose as a scenario of long distance homing to a submerged mothership with a resource constrained AUV and use as an example scenario to thoroughly describe the procedure of modeling and implementing a basic STERN solution for it.

Through simulations and experimental datasets, we provide an overview of the quantitative and qualitative performance and the flexibility associated to using factor graphs to develop a complete state estimation solution. We provide further insight on the requirements and details necessary for the deployment of such a solution on a real robotic platform, and discuss the shortcomings and limitations of our own implementation.

As future work, we look at a few different avenues open for research: (1) addressing the found limitations by removing the dynamic assumptions for the target's motion (i.e., constant heading, depth, and velocity) and generalizing the motion model with, for example, a Gaussian process factor; (2) extending the scenario to cover all prox-ops phases, i.e., carrying out the change of representation when crossing the hybrid-sensing threshold for full 6-DoF estimation of the target; (3) closing the loop on the remaining pipeline---i.e., to connect the output of the estimator to the planning and control modules onboard the chaser; and (4) further investigating the noise characteristics of USBL range and bearing measurements to develop more principled and robust methods for including these sparse, noisy, and outlier-prone relative measurements in our framework to achieve better real-time incremental performance. 

\section*{Acknowledgment}
We thank Ulysse Dhomé and everyone from the Naval Architecture department at KTH who helped with the logistics and practicalities involved in our field experiments.

\section*{Appendix}
\subsection{Motion Model Factor Jacobian}
\label{sec:mm_jacobian}
In Section \ref{sec:mm_factor} we established that the target's state consist of only a 3D position, and its dynamics are constrained via a motion model that assumes a constant depth, heading, and velocity. Because we developed our own motion model factor, for completion, we expand Equation~\eqref{eq:mm_residual} and derive the Jacobian $\bm J_{\bm{\epsilon}_{\text{mm}}}$ used by the nonlinear optimization routine for linearization. We compute the residual as
\begin{equation}
\bm \epsilon_{\text{mm}}(\tgtx{_i}, \tgtx{_j}, v, \theta) = 
    \begin{bmatrix}
       \tgtx{_i}{_x} + \cos(\theta)v\Delta t - \tgtx{_j}{_x} \\
        \tgtx{_i}{_y} + \sin(\theta)v\Delta t - \tgtx{_j}{_y} \\
        \tgtx{_i}{_z} - \tgtx{_j}{_z}
    \end{bmatrix},
\end{equation}

where $\tgtx{_i}$ and $\tgtx{_j}$ are consecutive positions of the target, $v$ and $\theta$ are the estimated velocity and heading, respectively, and $\Delta t$ is the time between discrete timestep $i$ and $j$.

We write the Jacobian matrix for the motion model factor as
\begin{equation}
    \bm J_{\bm{\epsilon}_{\text{mm}}} = [ \bm J_{{\bm{\epsilon}_{\text{mm}}}/\tgtx{_i}} \quad \bm J_{{\bm{\epsilon}_{\text{mm}}}/\tgtx{_j}} \quad \bm J_{{\bm{\epsilon}_{\text{mm}}}/v}\quad \bm J_{{\bm{\epsilon}_{\text{mm}}}/\theta}],
\end{equation}
with:
\begin{equation}
    \bm J_{{\bm{\epsilon}_{\text{mm}}}/\tgtx{_i}} = \bm I_{3\times3},
\end{equation}
\begin{equation}  
    \bm J_{{\bm{\epsilon}_{\text{mm}}}/\tgtx{_j}} = -\bm I_{3\times3},
\end{equation}
\begin{equation}
    \bm J_{{\bm{\epsilon}_{\text{mm}}}/v} = [\cos(\theta)\Delta t\quad \sin(\theta)\Delta t\quad 0]^T,
\end{equation}
and 
\begin{equation}
    \bm J_{{\bm{\epsilon}_{\text{mm}}}/\theta} = [-\sin(\theta) v \Delta t\quad \cos(\theta) v \Delta t \quad 0]^T.
\end{equation}

\subsection{Depth Factor Jacobian}
\label{sec:depthf_jacobian}
Similar to the motion model factor, for our SLAM implementation, we developed a depth factor that constrains the chaser's depth ($\chx{_z}$) with a single absolute measurement provided by the onboard pressure sensor. With the depth factor, we seek to constrain a single degree of freedom (Z position) from the chaser’s pose, however, because we are tracking the full joint 6DoF pose of the chaser $\chx$ which does not not belong to a vector space, but to $\SE(3)$ instead, in order to compute the incremental update $\bm{\hat{\xi}} = (\delta_\theta, \delta_\psi, \delta_\phi, \delta_x, \delta_y, \delta_z)$ used by the nonlinear optimization solver, we must use the exponential map of the updated variable. To describe this procedure, we follow closely the details presented in \href{https://gtsam.org/tutorials/intro.html#magicparlabel-65484}{Section 3.2 of the GTSAM tutorial}. The exponential map for a pose on $\SE(3)$ is written as 
\begin{equation}
    \text{exp}^{\bm{\hat{\xi}}} \approx 
    \begin{bmatrix}
        & & \delta_x\\
        & \Skew{\bm{\omega}}_{3\text{x}3} & \delta_y\\
        & & \delta_z\\
        & \bm{0}_{1\text{x}3} & 1
    \end{bmatrix},
\end{equation}
where $\Skew{\bm{\omega}}_{3\text{x}3}$ is the skew-symmetric matrix representing an incremental update for a rotation on $\text{SO}(3)$ defined by the vector $\bm{\omega} = [\delta_\theta, \delta_\psi, \delta_\phi]^T$ (yaw, pitch, roll). 

The observation model for the depth factor $h_{\text{d}}$ only maps the pose $\bm{q}$ from the navigation manifold $\chx \in \mathcal{X}$ to the scalar depth $q_z = \chx{_z}$:
\begin{equation}
     h_{\text{d}}(\bm{q}) = q_z.
\end{equation}

Defining the Jacobian of the depth factor as the matrix $\bm{J}_\text{d}$ such that 
\begin{equation}
    h_\text{d}(\bm{q} \text{exp}^{\bm{\hat{\xi}}}) \approx h_{\text{d}}(\bm{q}) + \bm{J}_\text{d}\bm{\xi},
\end{equation}
which, using the above equations, can be re-written as
\begin{equation}
\label{eq:depthf_obs_jacobian}
\begin{aligned}
     h_\text{d}(\bm{q} \text{exp}^{\bm{\hat{\xi}}}) &\approx h_\text{d}
     \left(
     \begin{bmatrix}
         & & q_x\\
        & \Rot{}{}{_{3\text{x}3}} & q_y\\
        & & q_z\\
        & \bm{0}_{1\text{x}3} & 1
     \end{bmatrix}
    \begin{bmatrix}
        & & \delta_x\\
        & \Skew{\bm{\omega}}_{3\text{x}3} & \delta_y\\
        & & \delta_z\\
        & \bm{0}_{1\text{x}3} & 1
    \end{bmatrix}
     \right) \\
     & \approx - \sin(q_\psi)\delta_x + \cos(q_\psi)\sin(q_\phi)\delta_y\\
     &\,\,\,\,\,+ \cos(q_\psi)\cos(q_\phi)\delta_z + q_z,
\end{aligned}
\end{equation}
where $\Rot{}{}{_{3\text{x}3}}$ is the 3 by 3 matrix representation of a rotation. We finally obtain $\bm{J}_\text{d}$ by taking the partial derivatives of Equation~\eqref{eq:depthf_obs_jacobian} as
\begin{equation}
    \bm{J}_\text{d} = \left[ \frac{\partial h_\text{d}}{\partial \delta_\theta} \quad \frac{\partial h_\text{d}}{\partial \delta_\psi} \quad \frac{\partial h_\text{d}}{\partial \delta_\phi} \quad \frac{\partial h_\text{d}}{\partial \delta_x} \quad \frac{\partial h_\text{d}}{\partial \delta_y} \quad \frac{\partial h_\text{d}}{\partial \delta_z} \right],
\end{equation}
resulting in
\begin{equation}
    \bm{J}_\text{d} = \left[0  \hspace{0.5em}  0  \hspace{0.5em}  0  \hspace{0.5em}  -\sin(q_\psi)  \hspace{0.5em}  \cos(q_\psi)\sin(q_\phi)  \hspace{0.5em}  \cos(q_\psi)\cos(q_\phi) \right].
\end{equation}

\subsection{Quantitative results for Section~\ref{sec:simulations}}
\begin{table}[h!]
\centering
\caption{Mean and standard deviation ($\sigma$) error results in $[\text{m/s}]$ used to plot the \textbf{velocity} error bars in Figure~\ref{fig:perpendicular_batch_results}.}
\label{tab:perpendicular_batch_velocity_errors}
\resizebox{0.9\linewidth}{!}{%
\begin{tabular}{@{}clllll@{}}
\toprule
\multicolumn{1}{l}{}                                                                                                                        & \multicolumn{5}{c}{\textbf{Filtered}}                                                                                                                                   \\ \midrule
\multicolumn{1}{c|}{\diagbox[width=\dimexpr \textwidth/18+2.5\tabcolsep\relax, height=0.8cm]{$\sigma_{\text{usbl}}$}{$\sigma_{\text{mm}}$}} & \multicolumn{1}{c}{\textbf{0}} & \multicolumn{1}{c}{\textbf{1}} & \multicolumn{1}{c}{\textbf{2}} & \multicolumn{1}{c}{\textbf{3}} & \multicolumn{1}{c|}{\textbf{4}}     \\ \midrule
\multicolumn{1}{c|}{\textbf{0}}                                                                                                             & 0.033$\pm$0.29                 & 0.026$\pm$0.15                 & 0.005$\pm$0.04                 & 0.015$\pm$0.01                 & \multicolumn{1}{l|}{\textbf{0.017$\pm$0.005}} \\
\multicolumn{1}{c|}{\textbf{1}}                                                                                                             & 0.072$\pm$0.26                 & 0.009$\pm$0.16                 & 0.016$\pm$0.05                 & 0.019$\pm$0.01                 & \multicolumn{1}{l|}{\hl{\textbf{0.018$\pm$0.001}}} \\
\multicolumn{1}{c|}{\textbf{2}}                                                                                                             & 0.031$\pm$0.24                 & 0.022$\pm$0.12                 & 0.011$\pm$0.04                 & 0.019$\pm$0.01                 & \multicolumn{1}{l|}{\textbf{0.027$\pm$0.005}} \\
\multicolumn{1}{c|}{\textbf{3}}                                                                                                             & 0.076$\pm$0.26                 & 0.022$\pm$0.14                 & 0.0134$\pm$0.051                 & \textbf{0.0477$\pm$0.016}                 & \multicolumn{1}{l|}{0.112$\pm$0.01} \\
\bottomrule
\end{tabular}%
}
\end{table}
\begin{table}[h!]
\centering
\caption{Mean and standard deviation ($\sigma$) errors in $[\text{deg}]$ used to plot the \textbf{heading} error bars in Figure~\ref{fig:perpendicular_batch_results}.}
\label{tab:perpendicular_batch_heading_errors}
\resizebox{0.9\linewidth}{!}{%
\begin{tabular}{@{}clllll@{}}
\toprule
\multicolumn{1}{l}{}                                                                                                                        & \multicolumn{5}{c}{\textbf{Filtered}}                                                                                                                                     \\ \midrule
\multicolumn{1}{c|}{\diagbox[width=\dimexpr \textwidth/18+2.5\tabcolsep\relax, height=0.8cm]{$\sigma_{\text{usbl}}$}{$\sigma_{\text{mm}}$}} & \multicolumn{1}{c}{\textbf{0}} & \multicolumn{1}{c}{\textbf{1}} & \multicolumn{1}{c}{\textbf{2}} & \multicolumn{1}{c}{\textbf{3}} & \multicolumn{1}{c|}{\textbf{4}}       \\ \midrule
\multicolumn{1}{c|}{\textbf{0}}                                                                                                             & -6.287$\pm$23.48               & -4.208$\pm$7.35                & -4.420$\pm$3.10                & -4.396$\pm$1.83                & \multicolumn{1}{l|}{\hl{\textbf{-4.666$\pm$0.70}}}  \\
\multicolumn{1}{c|}{\textbf{1}}                                                                                                             & -3.863$\pm$23.28               & -3.943$\pm$6.77                & -3.615$\pm$3.28                & -4.721$\pm$1.97                & \multicolumn{1}{l|}{\textbf{-4.957$\pm$0.72}}  \\
\multicolumn{1}{c|}{\textbf{2}}                                                                                                             & -3.476$\pm$19.85               & -6.003$\pm$7.66                & -4.367$\pm$3.22                & \textbf{-4.114$\pm$1.89}                & \multicolumn{1}{l|}{-5.217$\pm$0.86}  \\
\multicolumn{1}{c|}{\textbf{3}}                                                                                                             & -3.136$\pm$20.43               & -5.578$\pm$7.75                & -4.472$\pm$3.27                & -4.444$\pm$1.94                & \multicolumn{1}{l|}{\textbf{-5.398$\pm$0.77}}  \\
\bottomrule
\end{tabular}%
}
\end{table}
\begin{table}[h!]
\centering
\caption{Mean and standard deviation ($\sigma$) errors in $[\text{m}]$ used to plot the \textbf{position} error bars in Figure~\ref{fig:perpendicular_batch_results}.}
\label{tab:perpendicular_batch_position_errors}
\resizebox{0.9\linewidth}{!}{%
\begin{tabular}{@{}clllll@{}}
\toprule
\multicolumn{1}{l}{}                                                                                                                        & \multicolumn{5}{c}{\textbf{Filtered}}                                                                                                                                    \\ \midrule
\multicolumn{1}{c|}{\diagbox[width=\dimexpr \textwidth/18+2.5\tabcolsep\relax, height=0.8cm]{$\sigma_{\text{usbl}}$}{$\sigma_{\text{mm}}$}} & \multicolumn{1}{c}{\textbf{0}} & \multicolumn{1}{c}{\textbf{1}} & \multicolumn{1}{c}{\textbf{2}} & \multicolumn{1}{c}{\textbf{3}} & \multicolumn{1}{c|}{\textbf{4}}      \\ \midrule
\multicolumn{1}{c|}{\textbf{0}}                                                                                                             & \textbf{0.784$\pm$0.00}                 & \textbf{0.784$\pm$0.00}                 & \textbf{0.784$\pm$0.00}                 &\textbf{ 0.784$\pm$0.00}                 & \multicolumn{1}{l|}{\textbf{0.784$\pm$0.00}}  \\
\multicolumn{1}{c|}{\textbf{1}}                                                                                                             & 0.791$\pm$0.00                 & 0.791$\pm$0.00                 & 0.791$\pm$0.00                 & 0.791$\pm$0.00                 & \multicolumn{1}{l|}{\textbf{0.790$\pm$0.00}}  \\
\multicolumn{1}{c|}{\textbf{2}}                                                                                                             & 0.894$\pm$0.01                 & 0.891$\pm$0.01                 & 0.873$\pm$0.01                 &\textbf{ 0.848$\pm$0.01}                 & \multicolumn{1}{l|}{0.854$\pm$0.01}  \\
\multicolumn{1}{c|}{\textbf{3}}                                                                                                             & 1.170$\pm$0.03                 & 1.139$\pm$0.03                 & 1.046$\pm$0.03                 &\textbf{ 0.998$\pm$0.03}                 & \multicolumn{1}{l|}{1.123$\pm$0.03}  \\
\bottomrule
\end{tabular}%
}
\end{table}
\begin{table}[h]
\centering
\caption{Mean and standard deviation ($\sigma$) errors in $[\text{m/s}]$ used to plot the \textbf{velocity} error bars in Figure~\ref{fig:parallel_batch_results}.}
\label{tab:parallel_batch_velocity_errors}
\resizebox{0.9\linewidth}{!}{%
\begin{tabular}{@{}clllll@{}}
\toprule
\multicolumn{1}{l}{}                                                                                                                        & \multicolumn{5}{c}{\textbf{Filtered}}                                                                                                                                    \\ \midrule
\multicolumn{1}{c|}{\diagbox[width=\dimexpr \textwidth/18+2.5\tabcolsep\relax, height=0.8cm]{$\sigma_{\text{usbl}}$}{$\sigma_{\text{mm}}$}} & \multicolumn{1}{c}{\textbf{0}} & \multicolumn{1}{c}{\textbf{1}} & \multicolumn{1}{c}{\textbf{2}} & \multicolumn{1}{c}{\textbf{3}} & \multicolumn{1}{c|}{\textbf{4}}      \\ \midrule
\multicolumn{1}{c|}{\textbf{0}}                                                                                                             & 0.097$\pm$0.18                 & 0.038$\pm$0.05                 & 0.044$\pm$0.02                 & 0.043$\pm$0.005                 & \multicolumn{1}{l|}{\textbf{0.044$\pm$0.002}}  \\
\multicolumn{1}{c|}{\textbf{1}}                                                                                                             & 0.071$\pm$0.15                 & 0.045$\pm$0.05                 & 0.039$\pm$0.02                 & 0.042$\pm$0.004                 & \multicolumn{1}{l|}{\textbf{0.043$\pm$0.001}}  \\
\multicolumn{1}{c|}{\textbf{2}}                                                                                                             & 0.061$\pm$0.08                 & 0.046$\pm$0.05                 & 0.043$\pm$0.02                 & \textbf{0.040$\pm$0.004}                 & \multicolumn{1}{l|}{0.044$\pm$0.002}  \\
\multicolumn{1}{c|}{\textbf{3}}                                                                                                             & 0.076$\pm$0.15                 & 0.046$\pm$0.05                 & 0.043$\pm$0.02                 & \textbf{0.037$\pm$0.01}                 & \multicolumn{1}{l|}{0.044$\pm$0.00}  \\
\bottomrule
\end{tabular}%
}
\end{table}
\begin{table}[h!]
\centering
\caption{Mean and standard deviation ($\sigma$) errors in $[\text{deg}]$ used to plot the \textbf{heading} error bars in Figure~\ref{fig:parallel_batch_results}.}
\label{tab:parallel_batch_heading_errors}
\resizebox{0.9\linewidth}{!}{%
\begin{tabular}{@{}clllll@{}}
\toprule
\multicolumn{1}{l}{}                                                                                                                        & \multicolumn{5}{c}{\textbf{Filtered}}                                                                                                                                     \\ \midrule
\multicolumn{1}{c|}{\diagbox[width=\dimexpr \textwidth/18+2.5\tabcolsep\relax, height=0.8cm]{$\sigma_{\text{usbl}}$}{$\sigma_{\text{mm}}$}} & \multicolumn{1}{c}{\textbf{0}} & \multicolumn{1}{c}{\textbf{1}} & \multicolumn{1}{c}{\textbf{2}} & \multicolumn{1}{c}{\textbf{3}} & \multicolumn{1}{c|}{\textbf{4}}       \\ \midrule
\multicolumn{1}{c|}{\textbf{0}}                                                                                                             & -0.404$\pm$19.71               & -4.344$\pm$4.46                & -3.282$\pm$1.44                & \hl{\textbf{-2.884$\pm$1.10}}                & \multicolumn{1}{l|}{-3.653$\pm$0.47}  \\
\multicolumn{1}{c|}{\textbf{1}}                                                                                                             & -4.110$\pm$15.70               & -4.403$\pm$5.20                & -3.391$\pm$1.86                & \hl{\textbf{-3.462$\pm$1.03}}                & \multicolumn{1}{l|}{-4.081$\pm$0.55}  \\
\multicolumn{1}{c|}{\textbf{2}}                                                                                                             & -2.265$\pm$9.30                & -4.797$\pm$4.45                & -4.055$\pm$2.24                & -4.626$\pm$1.43                & \multicolumn{1}{l|}{\textbf{-5.124$\pm$0.66}}  \\
\multicolumn{1}{c|}{\textbf{3}}                                                                                                             & -1.943$\pm$16.69               & -5.240$\pm$5.51                & -4.777$\pm$2.19                & -6.065$\pm$1.55                & \multicolumn{1}{l|}{\hl{\textbf{-5.196$\pm$0.83}}}  \\
\bottomrule
\end{tabular}%
}
\end{table}
\begin{table}[h]
\centering
\caption{Mean and standard deviation ($\sigma$) errors in $[\text{m}]$ used to plot the \textbf{position} error bars in Figure~\ref{fig:parallel_batch_results}.}
\label{tab:parallel_batch_position_errors}
\resizebox{0.9\linewidth}{!}{%
\begin{tabular}{@{}clllll@{}}
\toprule
\multicolumn{1}{l}{}                                                                                                                        & \multicolumn{5}{c}{\textbf{Filtered}}                                                                                                                                    \\ \midrule
\multicolumn{1}{c|}{\diagbox[width=\dimexpr \textwidth/18+2.5\tabcolsep\relax, height=0.8cm]{$\sigma_{\text{usbl}}$}{$\sigma_{\text{mm}}$}} & \multicolumn{1}{c}{\textbf{0}} & \multicolumn{1}{c}{\textbf{1}} & \multicolumn{1}{c}{\textbf{2}} & \multicolumn{1}{c}{\textbf{3}} & \multicolumn{1}{c|}{\textbf{4}}      \\ \midrule
\multicolumn{1}{c|}{\textbf{0}}                                                                                                             & \textbf{0.564$\pm$0.00}                 & \textbf{0.564$\pm$0.00}                 & \textbf{0.564$\pm$0.00}                 & \textbf{0.564$\pm$0.00}                 & \multicolumn{1}{l|}{\textbf{0.564$\pm$0.00}}  \\
\multicolumn{1}{c|}{\textbf{1}}                                                                                                             & 0.591$\pm$0.00                 & 0.591$\pm$0.00                 & 0.589$\pm$0.00                 &\textbf{ 0.586$\pm$0.00}                 & \multicolumn{1}{l|}{0.588$\pm$0.00}  \\
\multicolumn{1}{c|}{\textbf{2}}                                                                                                             & 0.874$\pm$0.01                 & 0.853$\pm$0.02                 & \textbf{0.801$\pm$0.01}                 & 0.829$\pm$0.01                 & \multicolumn{1}{l|}{1.044$\pm$0.01}  \\
\multicolumn{1}{c|}{\textbf{3}}                                                                                                             & 1.398$\pm$0.03                 & 1.316$\pm$0.03                 & \textbf{1.159$\pm$0.03}                 & 1.261$\pm$0.02                 & \multicolumn{1}{l|}{1.731$\pm$0.03}  \\
\bottomrule
\end{tabular}%
}
\end{table}

\pagebreak
\subsection{Resulting IMU biases from the experimental deployments.}

\begin{table}[h!]
\centering
\caption{Final values for the estimated IMU biases in acceleration and angular rate per deployment. Time series is shown in Figure~\ref{fig:imu_biases}.}
\label{tab:imu_biases}
\resizebox{0.9\linewidth}{!}{%
\begin{tabular}{@{}cccccccccc@{}}
\toprule
                                     & \multicolumn{9}{c}{\textbf{Accelerometer bias [m/$s^2$]}}                                                          \\ \midrule
\multicolumn{1}{c|}{\textbf{test}}   & \textbf{1} & \textbf{2} & \textbf{3} & \textbf{4} & \textbf{5} & \textbf{6} & \textbf{7} & \textbf{8} & \textbf{9} \\ \midrule
\multicolumn{1}{c|}{\textbf{x-axis}} & -0.2171    & -0.2799    & -0.2442    & -0.4065    & -0.2224    & -0.3110    & -0.2872    & -0.0659    & -0.2679    \\
\multicolumn{1}{c|}{\textbf{y-axis}} & -0.1036    & -0.0067    & 0.0574     & -0.0475    & 0.0075     & 0.0264     & 0.0370     & 0.0156     & 0.2083     \\
\multicolumn{1}{c|}{\textbf{z-axis}} & 0.0250     & 0.0296     & 0.0273     & 0.0365     & 0.0220     & 0.0418     & 0.0269     & -0.0001    & 0.0308     \\ \midrule
\multicolumn{1}{l}{}                 & \multicolumn{9}{c}{\textbf{Gyroscope bias [rad/s]}}                                                                \\ \midrule
\multicolumn{1}{c|}{\textbf{test}}   & \textbf{1} & \textbf{2} & \textbf{3} & \textbf{4} & \textbf{5} & \textbf{6} & \textbf{7} & \textbf{8} & \textbf{9} \\ \midrule
\multicolumn{1}{c|}{\textbf{x-axis}} & -0.0012    & -0.0012    & -0.0017    & -0.0014    & -0.0014    & -0.0011    & -0.0012    & -0.0012    & -0.0015    \\
\multicolumn{1}{c|}{\textbf{y-axis}} & 0.0025     & 0.0025     & 0.0026     & 0.0026     & 0.0026     & 0.0025     & 0.0024     & 0.0025     & 0.0026     \\
\multicolumn{1}{c|}{\textbf{z-axis}} & -0.0032    & -0.0037    & -0.0076    & -0.0060    & -0.0060    & -0.0028    & -0.0037    & -0.0040    & -0.0054   
\end{tabular}%
}
\end{table}

\clearpage
\bibliographystyle{ieeetr}
\bibliography{references.bib}

\begin{IEEEbiography}[{\includegraphics[width=1in,height=1.25in,clip,keepaspectratio]{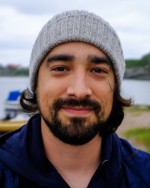}}]{Aldo Ter\'an Espinoza}
received the B.S. degree in automotive engineering from the Instituto Polit\'ecnico Nacional in Mexico City, Mexico, in 2017, and the M.Sc. degree in computer science and engineering from the KTH Royal Institute of Technology, Stockholm, Sweden, in 2020.

He is currently working towards a Ph.D degree in the Swedish Maritime Robotics Centre at the KTH Royal Institute of Technology in Stockholm, Sweden. He previously worked as a research engineer developing underwater sampling and perception systems. His research interests focus on probabilistic inference for underwater robotic perception and navigation. 
\end{IEEEbiography}

\begin{IEEEbiography}[{\includegraphics[width=1in,height=1.25in,clip,keepaspectratio]{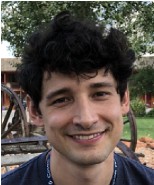}}]{Antonio Ter\'an Espinoza}
is a member of the Perception group at Waymo. He received his B.S. from the
Universidad Nacional Aut\'onoma de M\'exico (2015),
and his S.M. (2017) and Ph.D (2020) degrees in
aeronautics and astronautics from the Massachusetts
Institute of Technology as a Fulbright student and
CONACyT fellow. His research interests include all
kinds of robotics problems, mainly focusing on autonomous navigation, mapping, and state estimation.\end{IEEEbiography}

\begin{IEEEbiography}[{\includegraphics[width=1in,height=1.25in,clip,keepaspectratio]{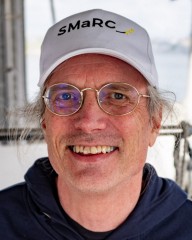}}]{John Folkesson}
(Senior Member, IEEE) received the
B.A. degree in physics from Queens College, City
University of New York, New York, NY, USA, in
1983, and the M.Sc. degree in computer science and
the Ph.D. degree in robotics from the Royal Institute
of Technology (KTH), Stockholm, Sweden, in 2001
and 2006, respectively.

He is currently a Professor of robotics
with the Robotics, Perception and Learning Lab, Center for Autonomous Systems, KTH. His research interests include navigation, mapping, perception, and
situation awareness for autonomous robots.\end{IEEEbiography}

\begin{IEEEbiography}[{\includegraphics[width=1in,height=1.25in,clip,keepaspectratio]{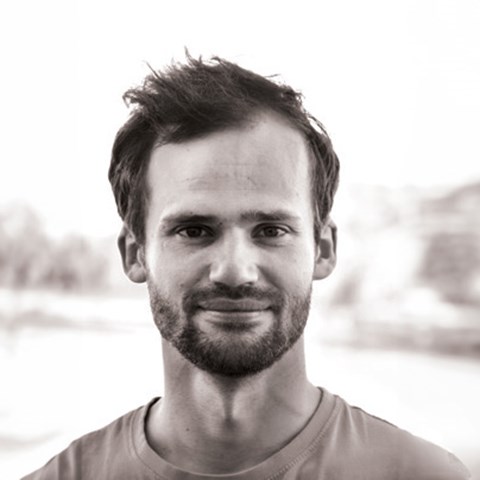}}]{Clemens Deutsch}
 received the B.Eng. degree from the Bremen University of Applied Sciences, Bremen, Germany, in 2015, and the M.Sc. degree in naval architecture as well as the Ph.D. degree in underwater robotics from the Royal Institute of Technology (KTH), Stockholm, Sweden, in 2017 and 2022, respectively. Currently, he is a postdoctoral researcher at the KTH Maritime Robotics Laboratory. His research focus lies on the design and performance optimization of underwater robots and other marine craft. In recent years, he has become an integral part in the development of the autonomous underwater vehicle LoLo.
\end{IEEEbiography}

\begin{IEEEbiography}[{\includegraphics[width=1in,height=1.25in,clip,keepaspectratio]{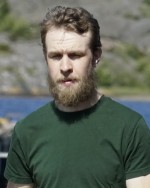}}]{Niklas Rolleberg}
received the B.S. degree in simulation technology and virtual design in 2016 from the KTH Royal Institute of Technology in Stockholm, Sweden. He is currently a research engineer with the Centre for Naval Architecture, leading the development of the AUV LoLo for the Swedish Maritime Robotics Centre, at the KTH Royal Institute of Technology in Stockholm, Sweden. His research interest include maritime vehicle development, environmental sampling technologies, and autonomy for underwater and surface vehicles.
\end{IEEEbiography}

\begin{IEEEbiography}[{\includegraphics[width=1in,height=1.25in,clip,keepaspectratio]{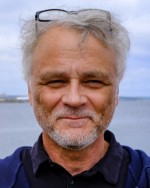}}]{Peter Sigray}
recieved his Ph.D degree in 1989 in atomic and molecular physics from the KTH Royal Institute of Technology in Stockholm, Sweden.

He has worked in the Swedish Defense Research Agency and as an adjunct lecturer and professor in the Department of Meteorology at Stockholm University. He is currently a researcher with the Centre for Naval Architecture at the KTH Royal Institute of Technology in Stockholm, Sweden. His current area of interest includes underwater acoustic communication for AUVs, underwater noise emission and bioacustics, and environmental monitoring.
\end{IEEEbiography}

\begin{IEEEbiography}[{\includegraphics[width=1in,height=1.25in,clip,keepaspectratio]{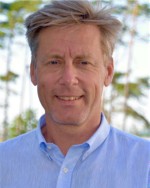}}]{Jakob Kuttenkeuler}
received the Ph.D. degree in
aeronautical engineering in 1998.

He is currently a Full Professor in naval architecture with the Centre for Naval Architecture, KTH Royal Institute of Technology, Stockholm, Sweden.
He is also a Senior Member with the Swedish Maritime Robotics Centre, KTH Royal Institute of Technology, where his research focuses on the fields of
autonomous underwater vehicles. He has dedicated
his research efforts into various fields ranging from
lightweight design, aircraft composites, aeroelastics,
robotic sailing, ocean monitoring, and maritime robotics.\end{IEEEbiography}

\end{document}